\documentclass[10pt,twocolumn,letterpaper]{article}

\usepackage{cvpr}              

\definecolor{cvprblue}{rgb}{0.21,0.49,0.74}
\usepackage[pagebackref,breaklinks,colorlinks,allcolors=cvprblue]{hyperref}

\usepackage{url}
\usepackage{caption}
\usepackage{wrapfig}
\usepackage{graphicx}
\usepackage{booktabs}
\usepackage{multirow}
\usepackage{enumitem}
\usepackage{algorithm}
\usepackage{algpseudocode}
\usepackage{ragged2e}
\usepackage[dvipsnames]{xcolor}
\usepackage[para]{footmisc}

\newcommand{\bmm}{\mathbf{m}}\newcommand{\bM}{\mathbf{M}}

\newcommand{\bp}{\mathbf{p}}

\newcommand{\br}{\mathbf{r}}

\newcommand{\bv}{\mathbf{v}}

\newcommand{\bx}{\mathbf{x}}

\newcommand{\cA}{\mathcal{A}}
\newcommand{\cB}{\mathcal{B}}
\newcommand{\cC}{\mathcal{C}}

\newcommand{\cS}{\mathcal{S}}

\makeatletter
\DeclareRobustCommand\onedot{\futurelet\@let@token\@onedot}
\def\@onedot{\ifx\@let@token.\else.\null\fi\xspace}

\makeatother

\definecolor{darkgreen}{rgb}{0,0.7,0}

\def\paperID{13882} 
\def\confName{CVPR}
\def\confYear{2025}

\title{SAMBLE: Shape-Specific Point Cloud Sampling for an Optimal \\ Trade-Off Between Local Detail and Global Uniformity}

\author{Chengzhi Wu$^{1}$ \quad\quad Yuxin Wan$^{1}$\thanks{Equal contribution.}  
\quad\quad Hao Fu$^{1}$\footnotemark[1]  \quad\quad Julius Pfrommer$^{2}$ \\
\quad Zeyun Zhong$^{1}$ \quad Junwei Zheng$^{1}$\thanks{Corresponding author.}
\quad Jiaming Zhang$^{1}$ \quad Jürgen Beyerer$^{1,2}$ 
\and
$^{1}$Karlsruhe Institute of Technology, Germany \quad\quad
$^{2}$Fraunhofer IOSB, Germany \vspace{-0.2cm}
\and
{\tt\footnotesize
\{chengzhi.wu, zeyun.zhong, junwei.zheng, jiangming.zhang\}@kit.edu, } \\
{\tt\footnotesize
\{yuxin.wan, hao.fu\}@student.kit.edu, \quad \{julius.pfrommer, juergen.beyerer\}@iosb.fraunhofer.de} 
\vspace{-0.1cm}}

\begin{document}
\maketitle
\begin{abstract}
Driven by the increasing demand for accurate and efficient representation of 3D data in various domains, point cloud sampling has emerged as a pivotal research topic in 3D computer vision. Recently, learning-to-sample methods have garnered growing interest from the community, particularly for their ability to be jointly trained with downstream tasks. However, previous learning-based sampling methods either lead to unrecognizable sampling patterns by generating a new point cloud or biased sampled results by focusing excessively on sharp edge details. Moreover, they all overlook the natural variations in point distribution across different shapes, applying a similar sampling strategy to all point clouds. In this paper, we propose a \textbf{S}parse \textbf{A}ttention \textbf{M}ap and \textbf{B}in-based \textbf{Le}arning method (termed SAMBLE) to learn shape-specific sampling strategies for point cloud shapes. SAMBLE effectively achieves an improved balance between sampling edge points for local details and preserving uniformity in the global shape, resulting in superior performance across multiple common point cloud downstream tasks, even in scenarios with few-point sampling. \vspace{-0.2cm}
\end{abstract}
\section{Introduction}
\label{sec:intro}
\begin{figure}[t]
\centering
\includegraphics[width=1\linewidth,trim=0 10 0 10,clip]{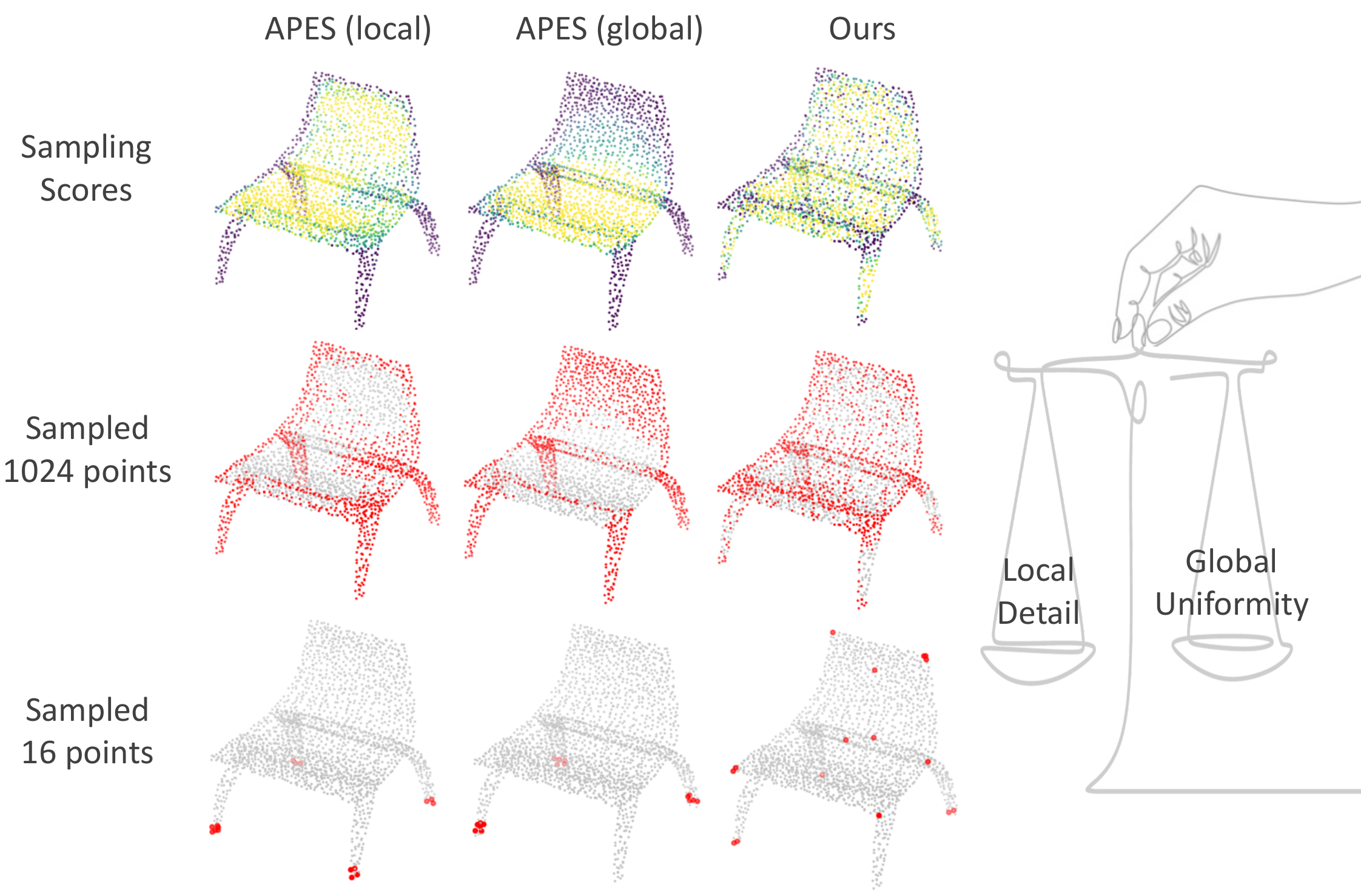}
\caption{Our method achieves an improved trade-off between sampling local details and preserving global uniformity. \vspace{-0.2cm}}
\label{fig:intro_tradeoff}
\end{figure}

Point cloud sampling is a less explored research area within the realm of this data representation. Traditional random sampling (RS) and farthest point sampling (FPS) remain the most commonly employed methods when sampling is required for point cloud learning. With the advancement of neural networks, several methods have emerged for point cloud sampling in a downstream task-oriented learning framework, including S-Net \cite{dovrat2019learning}, SampleNet \cite{lang2020samplenet}, MOPS-Net \cite{Qian2020MOPSNetAM}, etc. However, these methods first generate a new, smaller-sized point cloud as a proxy rather than directly sampling points from the original input, rendering the techniques akin to black-box neural network models with limited interpretability. Consequently, discerning geometric patterns in their qualitative results becomes challenging, as their outcomes closely resemble those obtained through random sampling. More recently, APES \cite{wu2023attention} pioneers the direction of using neural networks to learn point-wise sampling scores, with which it subsequently samples points whose scores are higher. However, with its score computation design and the Top-M sampling strategy, APES excessively focuses on local details of edge points, resulting in a deficiency in preserving good global uniformity of the input shapes. Consequently, the interpolation operation becomes impractical during the upsampling process, and the sampling quality of few-point sampling is notably subpar (see \cref{fig:intro_tradeoff}). In this paper, we introduce a novel point cloud sampling method that addresses the limitations of prior approaches, aiming to achieve a refined balance between capturing local details and preserving global uniformity. 

The concept originates from rethinking the mathematical characteristics of local details within point cloud shapes. Typically, these local details are represented by edge points that define the shape's outline and sharpest features. Is there a point property that can easily distinguish between different categories, such as edge points and non-edge points? The answer is affirmative. In our investigation, we have uncovered an extremely fundamental yet crucial observation: if point $\mathbf{p}_i$ is one of the $k$-nearest neighbors of point $\mathbf{p}_j$, it does not necessarily imply that $\mathbf{p}_j$ is also among the $k$-nearest neighbors of $\mathbf{p}_i$. Consequently, it leads to the conclusion that the \emph{frequency of each point being chosen as a neighbor} exhibits variation across a single point cloud. 

\begin{figure}[t]
\centering
\includegraphics[width=1\linewidth,trim=5 2 2 2,clip]{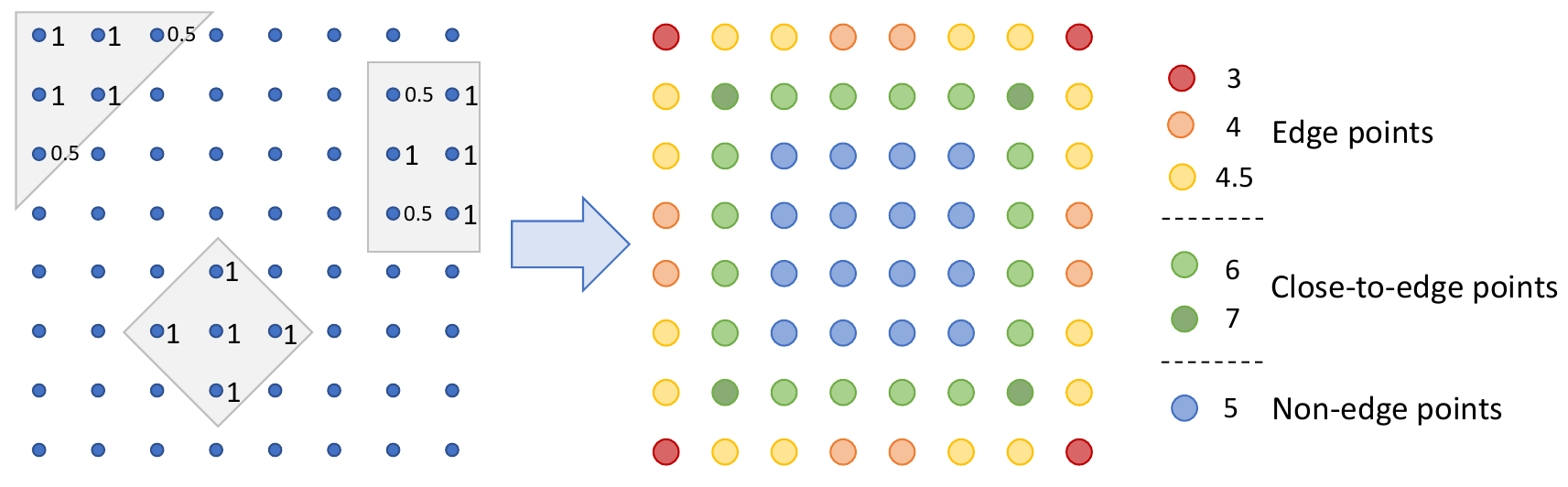}
\caption{When selecting an equal number of neighbors for each point in the input point cloud, points at different positions are chosen as neighbors with varying frequencies (numbers on the right). \vspace{-0.5cm} }
\label{fig:closeToEdge}
\end{figure}

We explore and demonstrate the importance of this point property with a simple example as illustrated in \cref{fig:closeToEdge}. Assume the input point cloud is a simple grid. When selecting 5 neighbors for each point, all three possible cases are given on the left (center point is self-contained as a neighbor). Note that in the triangular and rectangular cases, they each has a ``quantum-entangled'' twin point pair, in which two points share the possibility of being chosen as the neighbor. While an equal number of neighbors is selected for each point in the input point cloud, points at different positions are chosen as neighbors with varying frequencies, as listed on the right of \cref{fig:closeToEdge}. From it, we can observe that in addition to the edge point and non-edge point categories, there is also another noteworthy point category of close-to-edge points. Moreover, within each category, the points can be further grouped into more sub-categories. Overall, this point property effectively captures the local characteristics of a shape, especially for shape outline and sharp details. Building on it, we propose Sparse Attention Map (SAM) and introduce new methods for computing point-wise sampling scores to effectively balance the trade-off between local and global sampling. See more details in \cref{sec:method_indexingMode}.

On the other hand, after the point-wise sampling scores are computed, previous methods employ a Top-M sampling strategy for all point cloud shapes, which exacerbates the issue of oversampling edge points. We argue that the na\"{i}ve top-M sampling strategy may not be optimal across all point cloud shapes for downstream tasks. For example, sampling more non-edge points enhances global uniformity, while sampling more close-to-edge points ``thickens'' the edge, both of which can potentially improve the performance on downstream tasks \cite{wu2023attention}. To address this, we introduce a novel bin-based method to explore better sampling strategies shape-specifically by leveraging all point categories. This approach enables the sampling of points with smaller sampling scores, further optimizing the local-global trade-off. As a result, our method dynamically adjusts the sampling strategy for each shape, leading to a more tailored and interpretable sampling process for improved performance.

Our main contributions are summarized as follows: 
\begin{itemize}[itemsep=0pt,topsep=0pt,left=0pt]
\item We propose a sparse attention map that directly integrates shape local and global information at the attention map level for point cloud sampling, introducing multiple methods for computing point-wise sampling scores.
\item We present a novel approach for learning bin boundaries to partition points within individual shapes, enabling shape-specific sampling strategies by incorporating additional bin tokens during the attention computation.
\item Our method successfully achieves an improved trade-off between capturing local details and preserving global uniformity for the sampling process, resulting in enhanced performance both qualitatively and quantitatively.
\end{itemize}

\section{Related Work}
\label{sec:relatedWork}
\textbf{Point Cloud Sampling.}
Point cloud sampling is a key process in 3D data handling for simplifying high-resolution dense point clouds. Over the past decades, non-learning-based methods \cite{eldar1997farthest, moenning2003fast,groh2018flex} have predominantly been used for point cloud sampling. While Farthest Point Sampling (FPS) \cite{eldar1997farthest} is the most widely used one \cite{qi2017pointnet++,li2018pointcnn,wu2019pointconv,qian2022pointnext,zhao2021point}, Random Sampling (RS) has also been frequently adopted \cite{zhou2018voxelnet,qi2020p2b,groh2018flex}. More recently, learning-based sampling methods have shown superior performance with task-oriented training. S-Net \cite{dovrat2019learning} represents a pioneering work of generating new point coordinates from global representations, while SampleNet \cite{lang2020samplenet} introduces a soft projection operation for better point approximation. Following S-Net, multiple learning-based methods have been proposed \cite{lin2021net,wang2021pst,nezhadarya2020adaptive,wang2023lightn}. MOPS-Net \cite{Qian2020MOPSNetAM} learns a transformation matrix and multiplies it with the original point cloud to generate the sampled one. By employing the attention mechanism to learn point-wise sampling scores, APES \cite{wu2023attention} captures the edge points in the input point clouds with a strong focus.

\vspace{2pt}
\noindent\textbf{Deep Learning on Point Clouds.}
In contrast to the voxelization-based methods \cite{maturana2015voxnet,jiang2018pointsift,le2018pointgrid} and multi-view-based methods \cite{lawin2017deep,boulch2017unstructured,audebert2016semantic,tatarchenko2018tangent}, point-based methods deal directly with point clouds. The pioneer studies of PointNet \cite{qi2017pointnet} and PointNet++ \cite{qi2017pointnet++} tackle point clouds through point-wise Multi-Layer Perceptrons (MLPs) and max-pooling operations. Subsequently, other research shifts focus towards constructing more efficient building blocks for local feature extraction, such as convolution-based ones \cite{li2018pointcnn,lin2020fpconv,zhu2023point,ahn2022projection,wu2019pointconv,thomas2019kpconv,wu2023pointconvformer} and graph-based ones \cite{wang2019dynamic,simonovsky2017dynamic,chen2021gapointnet,zhang2021linked,xu2020grid,lin2020convolution,liu2019relation,lang2021dpc}. More recently, while MLP-based methods like PointNeXt \cite{qian2022pointnext} and PointMetaBase \cite{lin2023meta} have rekindled people's interest, the application of attention mechanisms to point cloud analysis has also garnered widespread attention \cite{vaswani2017attention,guo2021pct,zhao2021point,yu2022point,engel2021point,wen2023learnable,wu2024rethinking,wu2023sim2real}. For example, PT \cite{zhao2021point,wu2022point,wu2024point} series improve the model performance by introducing subtraction-based attention blocks, and \cite{wu2024rethinking} performs a large ablation study over attention module designs for point cloud processing. In addition, approaches that apply Transformers for point cloud self-supervised learning have also been proposed and explored \cite{yu2022point,pang2022masked,zhang2022point,wu2024cross,liu2022masked}.

\section{Methodology}
\label{sec:methodology}
\begin{figure}[t]
\centering
\includegraphics[width=\linewidth,trim=0 0 0 10,clip]{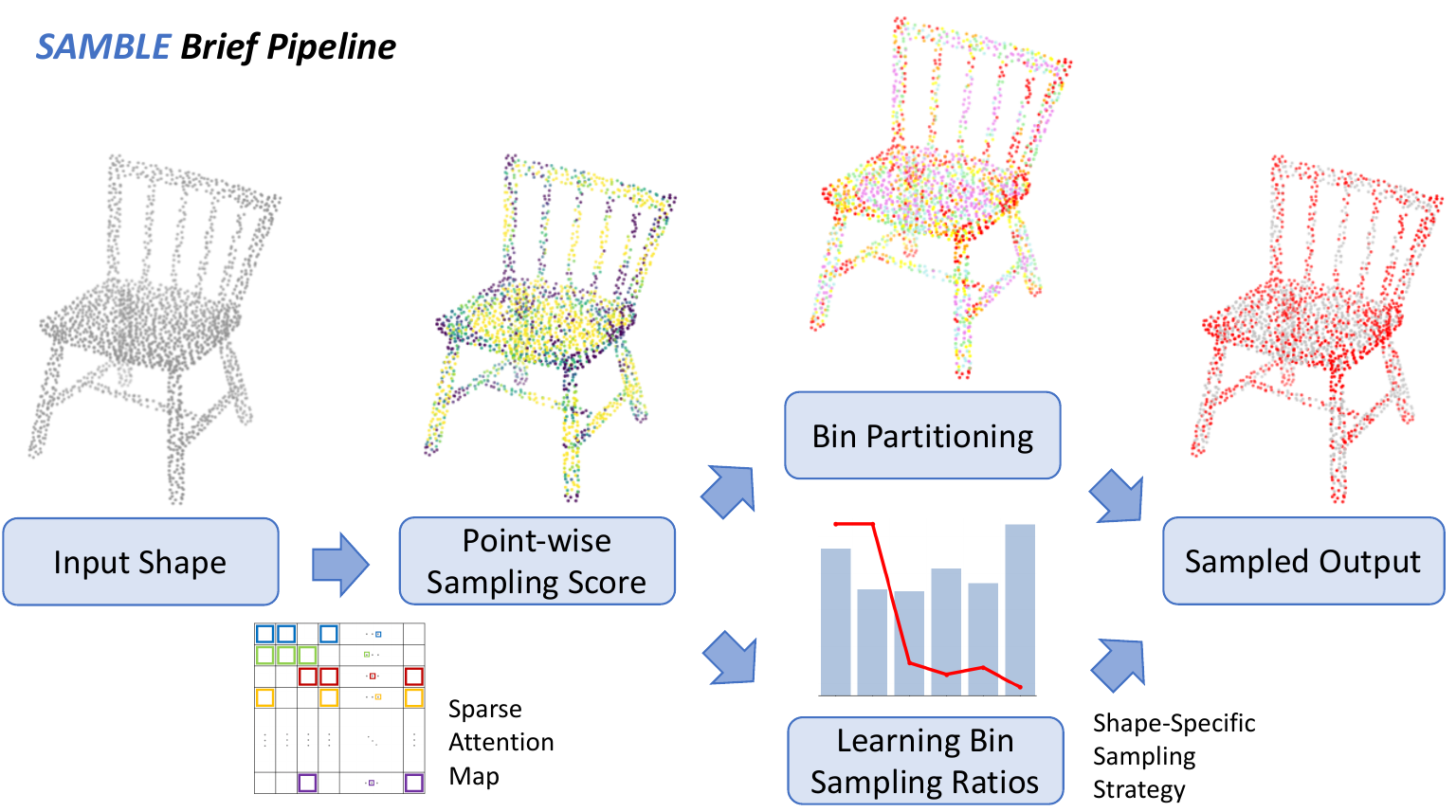}
\caption{A brief pipeline of our proposed method SAMBLE. It learns shape-specific sampling strategies for point cloud shapes.  \vspace{-0.2cm}}
\label{fig:SAMBLE_intro}
\end{figure}

A brief pipeline of SAMBLE is illustrated in \cref{fig:SAMBLE_intro}. It consists of three key steps: constructing a sparse attention map, computing point-wise sampling scores, and learning shape-specific sampling strategies through bin partitioning.

\subsection{Sparse Attention Map}
\label{sec:method_SAM}
\textbf{Local and Global Attention Maps.}
Both local and global attention maps are widely used in point cloud analysis. A global attention map is derived from the application of classical self-attention to point features of all points, while a local attention map concentrates on a point-centered area wherein cross-attention is specifically applied to the central point and its neighbors.

Denote $\cS_i$ as the set of $k$-nearest neighbors of point $\bp_i$, the local attention map for $\bp_i$ is defined as
\begin{equation}
\setlength{\abovedisplayskip}{5pt} 
\setlength{\belowdisplayskip}{5pt}
\bmm^l_i = \mathrm{softmax}\left( Q(\bp_i) K(\bp_{ij} - \bp_i)^{\top}_{j\in \cS_i}/\sqrt{d}\right) ,
\label{equ:local_attention_map}
\end{equation}
where $Q$ and $K$ stand for the linear layers applied on the query and key input, and the square root of the feature dimension count $\sqrt{d}$ serves as a scaling factor \cite{vaswani2017attention}. 

For the global attention map which is equivalent to taking all points as neighbors for each point, it is defined as 
\begin{equation}
\setlength{\abovedisplayskip}{5pt} 
\setlength{\belowdisplayskip}{5pt}
\bM^g = \mathrm{softmax}\left( Q(\bp_i) K(\bp_j)^{\top}_{i,j\in \cS}/\sqrt{d}\right) ,
\label{equ:global_attention_map}
\end{equation}
where $\cS$ denotes the set of all input points.

\begin{figure}[t]
\centering
\includegraphics[width=0.95\linewidth,trim=0 0 0 0,clip]{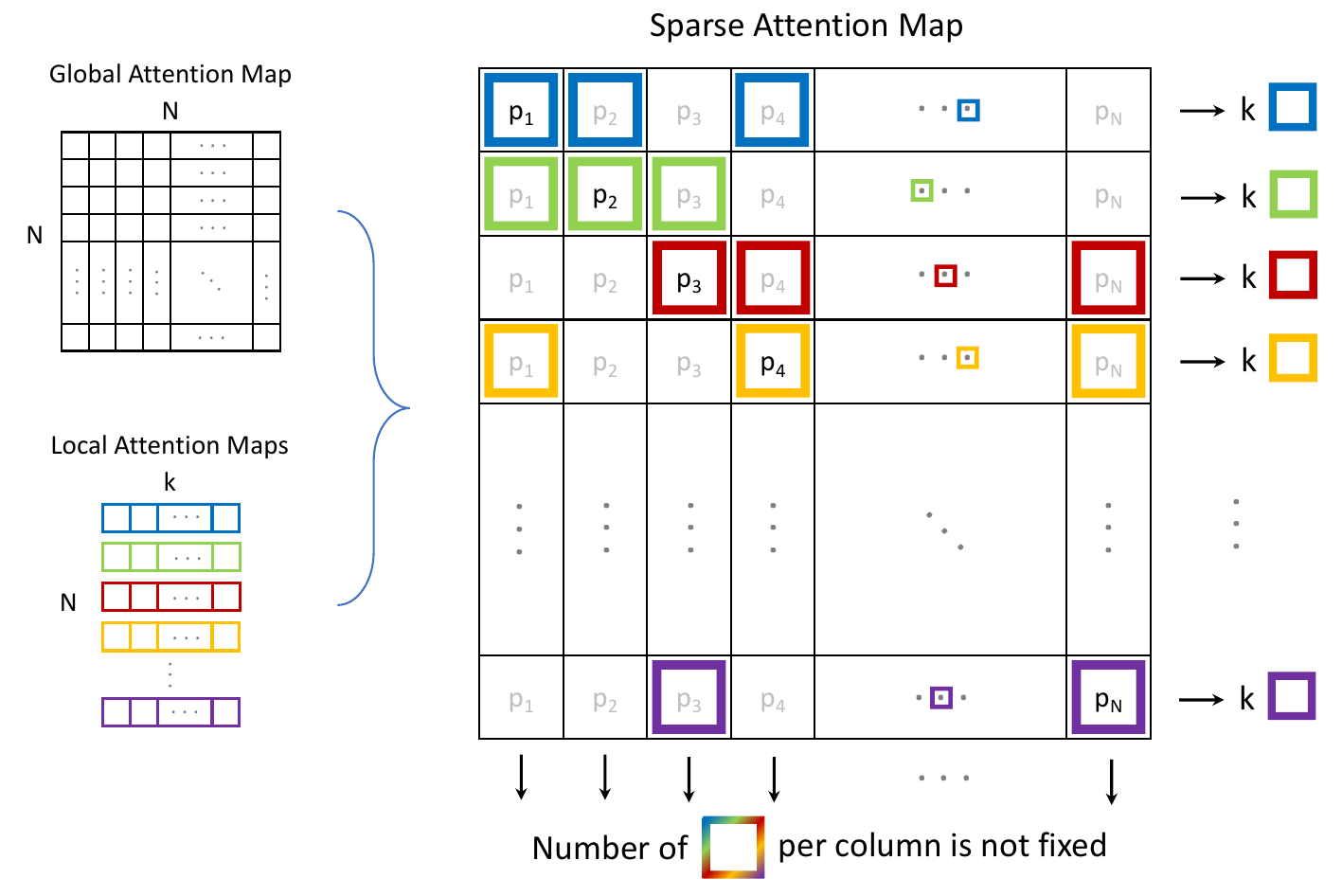}
\caption{Sparse attention map. In each row, $k$ cells are selected based on the KNN neighbor indexes for each point. The values of other non-selected cells are all set to 0. Note that the number of cells selected within each column is variable. \vspace{-0.2cm}}
\label{fig:SAM}
\end{figure}

\begin{table*}[h]
\centering
\resizebox{0.95\textwidth}{!}{
\begin{tabular}{@{}l|l|l|l@{}}
\toprule
Indexing Mode & Attention Map & Formula & Remark \\ \midrule
(i) Row standard deviation & Full & $a_{\bp_o} = f_{\text{std}}(\{m_{oj} | j = 1,2,\ldots,N \})$ & $f_{\text{std}}$: Computes standard deviation for a set of values \\
(ii) Column sum & Full & $a_{\bp_o} = \sum_{i=1}^{N} m_{io}$ &        \\ \midrule
(iii) Row standard deviation & Sparse &  $a_{\bp_o} = f_{\text{std}}(\{m^s_{oj} | j \in S_o \})$ &        $S_o$: Set of indices of selected cells in $o$th row\\
(iv) Row sum & Sparse & $a_{\bp_o} = \sum_{j=1}^{N} m^s_{oj}$ & Non-selected cells are all of $0$s   \\
(v) Column sum & Sparse & $a_{\bp_o} = \sum_{i=1}^{N} m^s_{io}$ &        \\
(vi) Column average & Sparse & $a_{\bp_o} = \sum_{i=1}^{N} m^s_{io} / n_o$ &  $n_o$: Number of selected cells in $o$th column \\
(vii) Column square-divided & Sparse & $a_{\bp_o} = \sum_{i=1}^{N} m^s_{io} / n_o^2$ & $n_o$: Number of selected cells in $o$th column \\ \bottomrule
\end{tabular}
}
\caption{Proposed different indexing modes for computing point-wise sampling scores.}
\label{tab:indexingModeFormula}
\end{table*}

\begin{figure*}[t]
\centering
\includegraphics[width=0.98\linewidth,trim=0 0 0 0,clip]{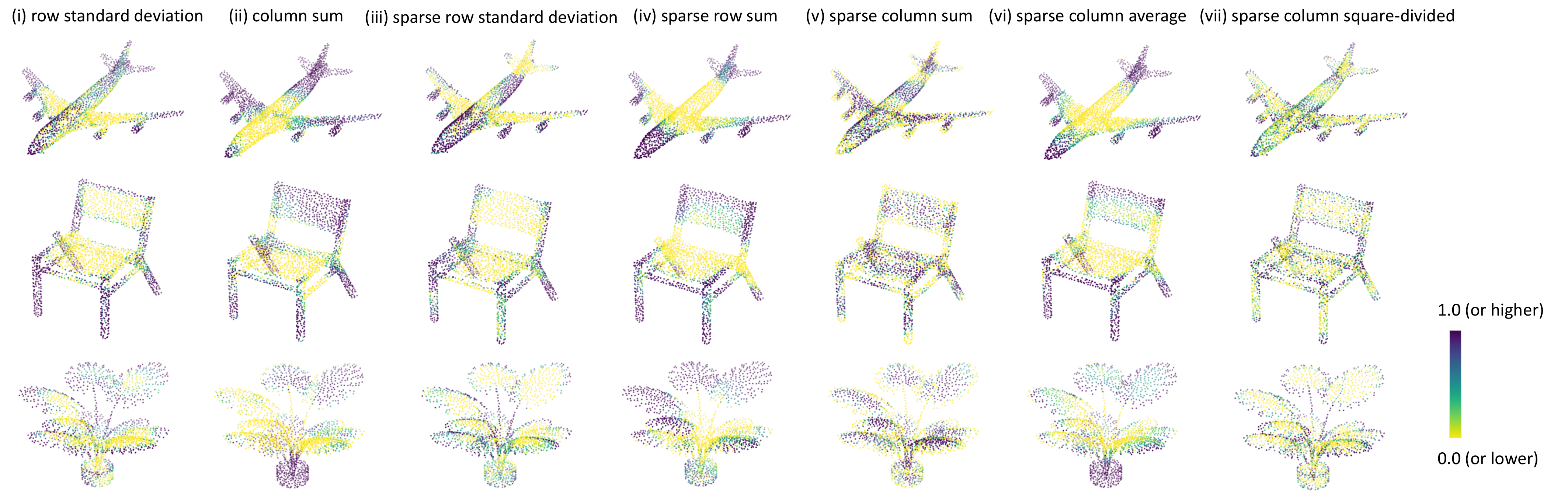}
\caption{Point sampling score heatmaps under different indexing modes. 
Scores are normalized to $\mathcal{N}(0.5,1)$ for better visualization.
\vspace{-0.2cm}}
\label{fig:indexingMode_carve}
\end{figure*}

\vspace{2pt}
\noindent\textbf{Sparse Attention Map.} 
Instead of using local or global attention maps solely, we propose sparse attention map, which combines the knowledge from both local and global information, to compute point-wise sampling scores. The idea is illustrated in \cref{fig:SAM}. After obtaining the global attention map with Eq. \ref{equ:global_attention_map}, $k$-NN is employed locally to find $k$ neighbors for each point. In this case, $k$ cells are being selected in each row. However, as discussed before, please notice that each point is chosen as a neighbor with varying frequencies. This means while for each row $k$ cells are selected, for each column, the number of selected cells varies. The selected cells are then ``carved" out to form the sparse attention map, with the values of other non-selected cells being set to 0. More vividly, consider the global attention map as a grid stone slab of size $N \times N$. For carve-based SAM, values of all cells are pre-computed and hidden in the slab grid cells, and only the selected cells are carved out.

\subsection{Computing Point-wise Sampling Score} 
\label{sec:method_indexingMode}
\textbf{Indexing Mode.}
When sampling points, the points are indexed based on the computed point-wise sampling scores. We call the method of computing point-wise sampling scores from the full/sparse attention map as Indexing Mode. With the original full attention map, following APES, there are two possible indexing modes: (i) row standard deviation; and (ii) column sum. For a global attention map $\bM^g$ of size $N \times N$, denote $m_{ij}$ as the value of $i$th row and $j$th column in $\bM^g$. These two indexing modes are formulated as modes (i) and (ii) in \cref{tab:indexingModeFormula}. To avoid possible confusion, we use notation $\bp_o$ to denote a point only in this subsection.

With the proposed sparse attention map, there are many other possible indexing modes. As discussed in \cref{sec:intro}, to achieve an improved trade-off between sampling edge points and preserving global uniformity, the frequency of each point being chosen as a neighbor, i.e., \emph{the number of selected cells in each column} is the key. The following indexing modes are designed and explored for comparison: (iii) sparse row standard deviation; (iv) sparse row sum; (v) sparse column sum; (vi) sparse column average; and (vii) sparse column square-divided. Again, for a sparse attention map $\bM^s$ of size $N \times N$, denote $m^s_{ij}$ as the value of $i$th row and $j$th column in $\bM^s$. For point $\bp_o$, we denote the set of indexes of the selected $k$ cells (indexes of $k$-nearest neighbors) in $o$th row as $S_o$, and denote the number of selected cells in $o$th column as $n_o$. Details and respective formulas of the proposed indexing modes are listed in \cref{tab:indexingModeFormula}.

\vspace{2pt}
\noindent\textbf{Heatmap.}
To analyze the behavior of each indexing mode, we train a separate model for each mode, ensuring that all other settings remain consistent. The sampling score distributions are visualized as heatmaps in \cref{fig:indexingMode_carve}, enhancing the interpretability of our method. From these heatmaps, we can see that both row-standard-deviation-based modes (modes i and iii) concentrate heavily on edge points. However, because they consistently prioritize thin or detailed regions, some areas may be overlooked. In contrast, modes ii and iv show less emphasis on edge points and instead distribute focus across a broader range of points, with a tendency toward other non-edge regions in a biased manner.

More interestingly, the comparison of modes v, vi, and vii, which utilize column-wise information from SAM, reveals distinct sampling preferences and strategies across different point categories. Mode v prioritizes non-edge points, mode vi emphasizes the global shape, and mode vii focuses slightly more on edge points. This is because edge points typically have a smaller number of $n_o$. Despite these differences and unique characteristics, all three modes capture the overall shape more uniformly compared to the former four. In our case, we aim to sample edge points without over-emphasizing them. For instance, when sampling detailed areas like chair legs, we want to capture some edge points without selecting them all, while also ensuring that non-edge points are sampled to preserve better global uniformity. Given this balance, we chose mode vii as the primary indexing mode for most of the experiments in the following sections. The detailed ablation study over different indexing modes is presented in \cref{sec:ablation}.

\subsection{Sampling with Bins}
\label{sec:method_bins}
After point-wise sampling scores are computed with SAM, points are sampled according to certain rules. The simplest approach is top-M sampling, where points with the highest scores are sampled. In our case, as we aim to enhance the local-global trade-off and leverage all point categories during the sampling process, we suggest employing a bin-based sampling strategy to allow for the possible sampling of certain close-to-edge points or even non-edge points.

\vspace{2pt}
\noindent\textbf{Bin Partitioning.}
The process begins with processing the distribution of normalized point-wise sampling scores $a_{\bp_i}$ across the shapes within the current batch. Let $n_b$ represent the number of bins used for partitioning, from which $n_b - 1$ bin boundary values need to be derived from the distribution. During each training step, a vector $\boldsymbol{\nu}_c = (\nu_1, \nu_2, \cdots, \nu_{n_b-1})$ is computed based on the point score distribution, ensuring an equitable division of points across all shapes in the current batch. Note that while $\boldsymbol{\nu}_c$ facilitates an even division at the batch level, the points within each individual shape are not necessarily evenly partitioned according to the batch-based bin boundary values.

During the training, for the first iteration, we directly use the boundary values derived from the first batch of data as the dynamic boundary values. Subsequently, since the second iteration, boundaries are updated adaptively in a momentum-based manner:
\begin{equation}
\setlength{\abovedisplayskip}{5pt} 
\setlength{\belowdisplayskip}{5pt}
\label{equ: dynamic bin boundaries}
    \boldsymbol{\nu}_t = \gamma \boldsymbol{\nu}_{t-1} + (1-\gamma)\boldsymbol{\nu}_c \, ,
\end{equation}
where $\boldsymbol{\nu}_{t-1}$ stands for the bin partitioning boundaries used in the last iteration, and $\boldsymbol{\nu}_t$ is the updated dynamic boundaries used for the current iteration. $\gamma \in (0, 1)$ is the momentum update factor. With updated boundary values $\boldsymbol{\nu}_t$, points in each shape are divided into $n_b$ subsets of $\{\cB_1, \cB_2, \ldots, \cB_{n_b}\}$ based on their sampling scores.

The core idea presented here is that, instead of using pre-engineered bin boundary values, we employ adaptive learning for bin boundaries, and they are gradually learned from the entire training dataset. These values are intended to evenly partition the distribution of point-wise sampling scores across all shapes and points in the training dataset. Consequently, for each individual shape, the acquired boundary values can effectively partition its points into bins with a shape-specific strategy, capturing the unique characteristics of the shape while maintaining a degree of proximity to other shapes within the dataset.

\vspace{2pt}
\noindent\textbf{Tokens for Learning Bin Weights.} 
With points already being partitioned into bins for each shape, the next step is to learn a shape-specific sampling strategy, i.e., to learn shape-specific sampling weights for each bin. Inspired by ViT \cite{dosovitskiy2020image}, VilT \cite{kim2021vilt}, and Mask3D \cite{schult2023mask3d} — which leverage additional tokens during the computation of attention maps to extract and convey information across the entire feature map or specific groups of points or pixels — we introduce additional tokens specifically for learning bin sampling weights. In our case, attention maps are computed shape-specific during the downsampling process, facilitating the learning of bin sampling weights also in a shape-specific manner.

\begin{figure}[t]
\centering
\includegraphics[width=1.0\linewidth,trim=0 10 0 10,clip]{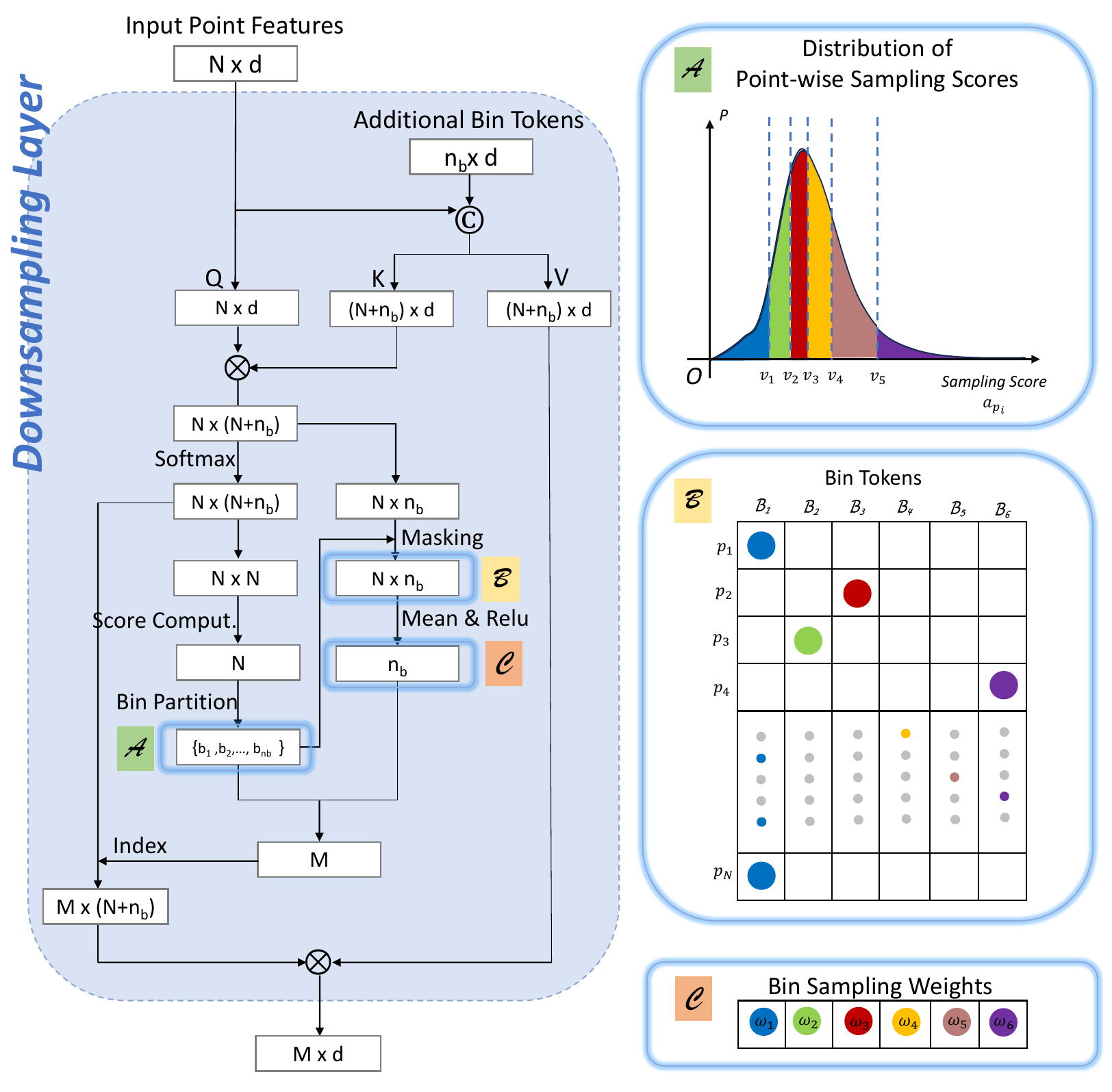}
\caption{Network structure of our proposed downsampling layer. Block $\cA$: Points in each shape are partitioned into $n_b$ bins.  Block $\cB$: Masking the split-out point-to-token sub-attention map. Block $\cC$: Learned bin sampling weights. \vspace{-0.2cm}} 
\label{fig:bin_token}
\end{figure}

Using the former proposed bin partitioning method, points in each shape are partitioned into $n_b$ subsets of $\{\cB_1, \cB_2, \ldots, \cB_{n_b}\}$. The sampling weight $\omega_{j}$ for bin $\cB_j (j=1, 2, \ldots, n_b)$ is established based on the distinctive features of each shape. \cref{fig:bin_token} gives the network structure of our proposed downsampling layer and illustrates the idea of using additional tokens. $n_b$ bin tokens are introduced during the attention computation, where each token corresponds to a specific bin. As shown in \cref{fig:bin_token}, the bin tokens are initially concatenated with the input point-wise features for $\mathit{Key}$ and $\mathit{Value}$. Subsequently, the combined features are subjected to a cross-attention mechanism with the original point-wise features as $\mathit{Query}$. The attention map is split into two parts of a point-to-point sub-attention map and a point-to-token sub-attention map. For the point-to-point attention map, the methods proposed in \cref{sec:method_SAM} and \cref{sec:method_indexingMode} are applied to it to obtain point-wise sampling scores. Note that in this case, the row-wise sum is not exactly equal to 1 but still very close to 1 since $n_b$ is of a very small quantity compared to $N$. With computed point scores, dynamic boundary values $\bv_t$ are obtained for bin partitioning. Using the information regarding the allocation of points to respective bins, a mask operation is performed on the point-to-token sub-attention map as illustrated in Block B of \cref{fig:bin_token}. 
The sampling weights $\omega_{j}$ are then subsequently acquired with
\begin{equation}
\setlength{\abovedisplayskip}{5pt} 
\setlength{\belowdisplayskip}{5pt}
\label{equ:token_to_sampling_weight}
    \omega_{j} = \text{ReLU}\left( \frac{1}{\beta_j}\sum_{\bp_i \in \cB_j} m_{\bp_i,\cB_j} \right) \, ,
\end{equation}
where $\beta_j$ stands for the number of points in bin $\cB_j$, and $m_{\bp_i,\cB_j}$ represents the element in the energy matrix corresponding to point $\bp_i$ in row and $\cB_j$ in column.

\begin{figure*}[t]
    \centering
    \includegraphics[width=1\linewidth,trim=0 0 0 0,clip]{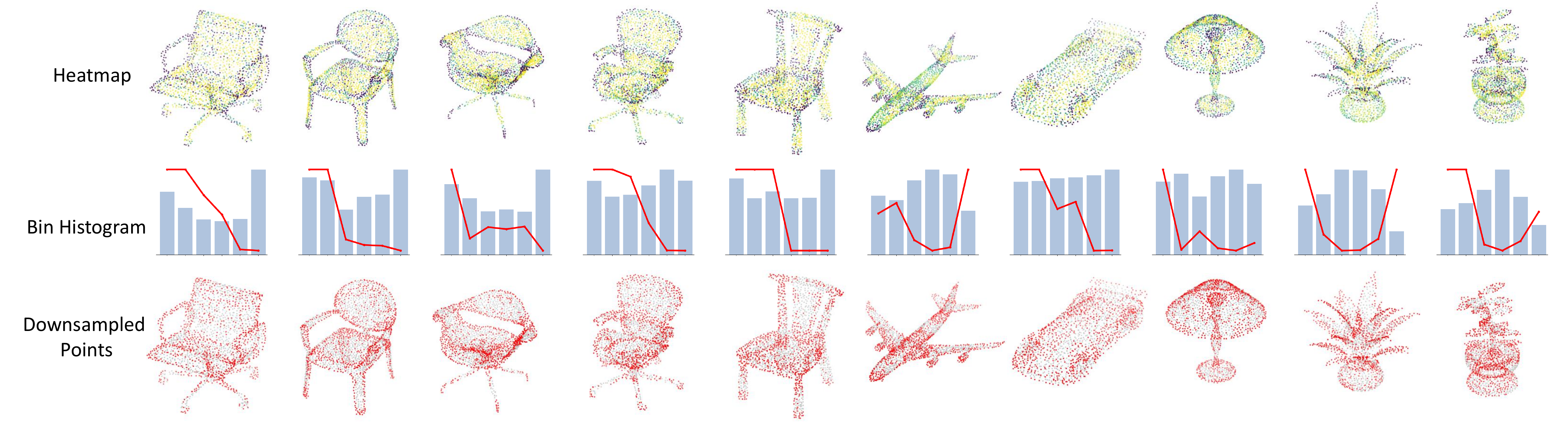}
    \caption{Qualitative results of our proposed SAMBLE. Apart from the sampled results, sampling score heatmaps and bin histograms along with bin sampling ratios are also given. All shapes are from the test set. Zoom in for optimal visual clarity.\vspace{-0.2cm}}
    \label{fig:cls}
\end{figure*}

\vspace{2pt}
\noindent\textbf{In-Bin Point Sampling.}
For each shape, by considering the number of points contained within bins $\boldsymbol{\beta} = (\beta_1, \beta_2, \ldots, \beta_{n_b})$ alongside the determined bin sampling weights $\boldsymbol{\omega} = (\omega_1, \omega_2, \ldots, \omega_{n_b})$, the specific numbers of points to be sampled from each bin $\boldsymbol{\kappa} = (\kappa_1, \kappa_2, \ldots,  \kappa_{n_b})$ need to be determined. Direct multiplication of $\boldsymbol{\beta}$ and $\boldsymbol{\omega}$ does not yield a sum that aligns with the total number of down-sampled points $M$ required by the network structure. To address this discrepancy, a scaling method is applied to first scale bin sampling weights $\omega_j$. Furthermore, to prevent $\kappa_j$ from surpassing the available point number $\beta_j$ in any bin, any excess points are proportionately redistributed to other bins that have not been fully sampled. The detailed algorithm is presented in the supplementary materials.

Finally, within bin $\cB_j$, $\kappa_j$ points are selected through random sampling with priors. The sampling probability $\rho_{\bp_i}$ is obtained via softmax operation over the normalized point sampling score $a_{\bp_i}$ with a temperature parameter $\tau$:
\begin{equation}
\label{equ: softmax}
\setlength{\abovedisplayskip}{5pt} 
\setlength{\belowdisplayskip}{3pt}
\rho_{\bp_i} = \frac{e^{a_{\bp_i}/\tau}}{\sum_{\bp_i \in \cB_j} e^{a_{\bp_i}/\tau}} \,.
\end{equation}

\section{Experiments}
\label{sec:experiments}
Like most related works, such as S-NET \cite{dovrat2019learning}, SampleNet \cite{lang2020samplenet}, and APES \cite{wu2023attention}, SAMBLE is specifically designed for point cloud shapes. To ensure a fair comparison, we conduct experiments using the same base network architecture as APES \cite{wu2023attention} on standard point cloud shape datasets, including ModelNet40 and ShapeNet-Part. 
It is important to note that point cloud sampling is not a standalone task; its effectiveness must be validated through downstream tasks.

\subsection{Classification}
\textbf{Experiment Setting.}
ModelNet40 classification benchmark \cite{wu20153d} includes 12,311 CAD models across 40 categories. For a fair comparison, we use the official train-test split, with 9,843 for training and 2,468 for testing. Points are uniformly sampled from the mesh surface and normalized to the unit sphere. Only 3D coordinates are used as input, with random scaling, rotation, and shifting applied for data augmentation. We use $n_b = 6$ bins for point partitioning. The momentum update factor $\gamma = 0.99$ for updating bin boundary values. The temperature parameter $\tau = 0.1$.

\begin{table}[t]
\centering
\resizebox{0.9\columnwidth}{!}{
\begin{tabular}{@{}lccc@{}}
\toprule
\multirow{2}{*}{Method} & \multirow{2}{*}{\begin{tabular}[c]{@{}c@{}} \rule{0pt}{2.5ex} Cls. \\ OA (\%) \end{tabular}} & \multicolumn{2}{c}{Seg.} \\ \cmidrule(l){3-4} 
 &  & Cat. mIoU (\%) & Ins. mIoU (\%) \\ \midrule
PointNet \cite{qi2017pointnet} & 89.2 & 80.4 & 83.7 \\
PointNet++ \cite{qi2017pointnet++} & 91.9 & 81.9 & 85.1 \\
SpiderCNN \cite{xu2018spidercnn} & 92.4 & 82.4 & 85.3 \\
DGCNN \cite{wang2019dynamic} & 92.9 & 82.3 & 85.2 \\
PointConv \cite{wu2019pointconv} & 92.5 & 82.8 & 85.7 \\
PT$^1$ \cite{engel2021point} & 92.8 & - & 85.9 \\
PT$^2$ \cite{zhao2021point} & 93.7 & 83.7 & 86.6 \\
PCT \cite{guo2021pct} & 93.2 & - & 86.4 \\
PRA-Net \cite{cheng2021net} & 93.7 & 83.7 & 86.3 \\
CurveNet \cite{muzahid2020curvenet} & 93.8 & - & 86.6 \\
DeltaConv \cite{wiersma2022deltaconv} & 93.8 & - & 86.6 \\ 
PointNeXt \cite{qian2022pointnext}  & 93.2 & 84.4 & \textbf{86.7} \\
PointMetaBase \cite{lin2023meta} & - & 84.3 & \textbf{86.7} \\
APES (local) \cite{wu2023attention} & 93.5 & 83.1 & 85.6 \\
APES (global) \cite{wu2023attention} & 93.8 & 83.7 & 85.8 \\ \midrule
SAMBLE & \textbf{94.2} & \textbf{84.5} & \textbf{86.7} \\ 
\bottomrule
\end{tabular}
} 
\caption{Classification and segmentation results on the ModelNet40 and ShapeNet-Part benchmarks. In comparison with other SOTA methods that also only use raw point cloud data as input. \vspace{-0.4cm}
}
\label{tab:cls_seg}
\end{table}

\vspace{2pt}
\noindent\textbf{Qualitative and Quantitative Results.}
Qualitative results of SAMBLE are presented in \cref{fig:cls}, including sampling score heatmaps, learned bin partitioning strategy with bin sampling ratios, and final sampled results. From it, we can observe that SAMBLE effectively samples sufficient edge points to capture the shape structure. Moreover, it maintains better global uniformity by avoiding excessive focus on edge points, particularly in those thin or sharp regions like chair legs. Logged shape bin histograms confirm the learning of shape-specific sampling strategies. More visualization results are provided in the supplementary materials, highlighting an intriguing pattern where shapes of the same category exhibit similar histogram distributions and sampling strategies. Overall, SAMBLE successfully achieves an improved trade-off between sampling edge points and preserving shape global uniformity. Quantitative results are provided in \cref{tab:cls_seg}. Our method performs better than previous approaches and achieves state-of-the-art performance.

\begin{figure}[t]
\centering
    \includegraphics[width=1\linewidth,trim=0 0 0 0,clip]{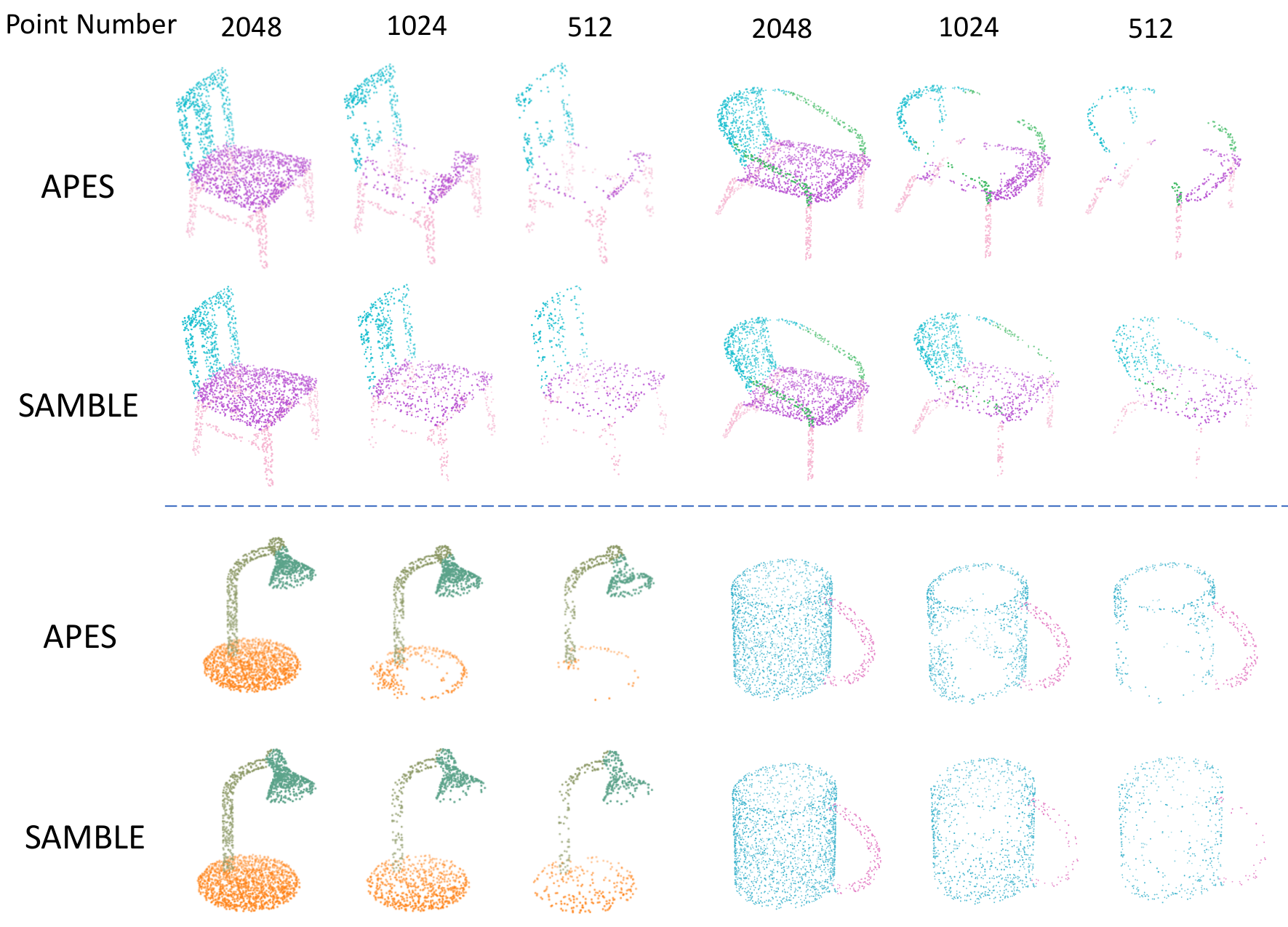}
    \captionof{figure}{Segmentation results of our proposed SAMBLE in comparison with APES. All shapes are from the test set.}
    \label{fig:seg}
\end{figure}

\begin{table}[t]
\centering
\resizebox{0.9\linewidth}{!}{
\begin{tabular}{@{}cccccccc@{}}
\toprule
Method         & \multicolumn{3}{c}{PointNeXt} &  & \multicolumn{3}{c}{SAMBLE} \\ \cmidrule(lr){2-4} \cmidrule(l){6-8} 
Point Number   & 2048   & 1024  & 512     &  & 2048  & 1024  & 512    \\ \midrule
Cat. mIoU (\%) & 84.40  & 83.79 & 82.77   &  & 84.51 & 84.84 & \textbf{85.04}    \\
Ins. mIoU (\%) & 86.70  & 86.18 & 85.18   &  & 86.67 & 86.93 & \textbf{87.12}    \\ \bottomrule
\end{tabular}
}
\caption{Additional segmentation performances evaluated on the intermediate downsampled sub-point clouds. \vspace{-0.05cm}}
\label{tab:samble_segInter}
\end{table}

\subsection{Segmentation}
\textbf{Experiment setting.}
The ShapeNet-Part dataset \cite{yi2016scalable} is used for 3D object part segmentation. It includes 16,880 3D models across 16 categories, with 14,006 models for training and 2,874 for testing. Each category contains 2–6 parts, totaling 50 distinct parts. We use the sampled point sets produced in \cite{qi2017pointnet} for a fair comparison with prior work. For evaluation metrics, we report both category mIoU and instance mIoU. We use $n_b = 4$ bins for point partitioning. The momentum update factor $\gamma = 0.99$ for updating bin boundary values. The temperature parameter $\tau = 0.1$. 

\vspace{2pt}
\noindent\textbf{Qualitative and Quantitative Results.}
Qualitative results are presented in \cref{fig:seg}, where we observe that, compared to APES, which tends to focus excessively on edge points, SAMBLE achieves a significantly improved balance between sampling edge points and preserving shape global uniformity. For example, SAMBLE demonstrates a more balanced use of non-edge points, as seen in the chair seat and lamp base, reflecting a thoughtful sampling strategy that accounts for different point categories and provides a more comprehensive representation of the overall shape. The quantitative results in \cref{tab:cls_seg} further highlight that SAMBLE achieves state-of-the-art performance.

For the part segmentation benchmark, we further report the performance on the intermediate downsampled sub-point clouds in \cref{tab:samble_segInter}. Additionally, results from PointNeXt \cite{qian2022pointnext} are also presented, which is a prominent point cloud learning method that employs FPS for downsampling. It is evident that FPS-based methods exhibit poorer performance when evaluated on intermediate downsampled sub-point clouds. In contrast, a notable observation is that SAMBLE achieves superior performance on intermediate downsampled points. This indicates that the learned sampled points contribute more to the overall performance, while the upsampling layer cannot fully reconstruct the features of the discarded points. Although SAMBLE enables the interpolation-based upsampling and it outperforms the upsampling layer used in APES (see details in \cref{sec:ablation}), there remains potential for further improvement by designing a more meticulously crafted upsampling layer. However, this is another topic and beyond the scope of this paper.

\begin{table*}[t]
\centering
\resizebox{\linewidth}{!}{
\begin{tabular}{c|cccccccccccc}
\toprule
$M$ \, &\, Voxel & RS &  FPS \cite{eldar1997farthest}  &  S-NET  \cite{dovrat2019learning}  &  PST-NET  \cite{wang2021pst}  & SampleNet \cite{lang2020samplenet} &  MOPS-Net  \cite{Qian2020MOPSNetAM}  &  DA-Net \cite{lin2021net}  &  LighTN \cite{wang2023lightn}  & \begin{tabular}[c]{@{}c@{}} APES \cite{wu2023attention} \\ (w/ pre-pro.) \end{tabular} & \begin{tabular}[c]{@{}c@{}} APES \cite{wu2023attention} \\ (w/o pre-pro.) \end{tabular} & SAMBLE \\ \midrule 
512 & 73.82 & 87.52 & 88.34 & 87.80 & 87.94 & 88.16 & 86.67 & 89.01 & 89.91 & \textbf{90.81} & 89.81 & 90.58 \\
256 & 73.50 & 77.09 & 83.64 & 82.38 & 83.15 & 84.27 & 86.63 & 86.24 & 88.21 & \textbf{90.40} & 86.78 & 90.18\\
128 & 68.15 & 56.44 & 70.34 & 77.53 & 80.11 & 80.75 & 86.06 & 85.67 & 86.26 & 89.77 & 84.87 & \textbf{90.02} \\
64 & 58.31 & 31.69 & 46.42 & 70.45 & 76.06 & 79.86 & 85.25 & 85.55 & 86.51 & 89.57 & 79.23 & \textbf{89.81} \\
32 & 20.02 & 16.35 & 26.58 & 60.70 & 63.92 & 77.31 & 84.28 & 85.11 & 86.18 & 88.56 & 75.63 & \textbf{89.45} \\ 
\bottomrule
\end{tabular}}
\caption{Comparison with other sampling methods. Evaluated on the ModelNet40 classification benchmark with multiple sampling sizes. For APES, we additionally report its performance when pre-processing is not performed for a fair comparison. \vspace{-0.1cm}}
\label{tab:compare}
\end{table*}

\subsection{Few-Point Sampling}
\textbf{Experiment setting.}
We further compare our sampling method to previous approaches, including RS, FPS, and more recent learning-based methods such as S-Net, SampleNet, LightTN, APES, and others. The evaluation follows the same framework as in \cite{dovrat2019learning, wang2023lightn, wu2023attention}. First, the point cloud is downsampled to a limited number of points, and the resulting subset is then fed into a task network for evaluation. For this comparison, we use the ModelNet40 classification task with the vanilla PointNet network. All sampling methods are evaluated across multiple sampling sizes of $M$.

\begin{figure}[t]
    \centering
    \includegraphics[width=1\linewidth,trim=0 0 0 0,clip]{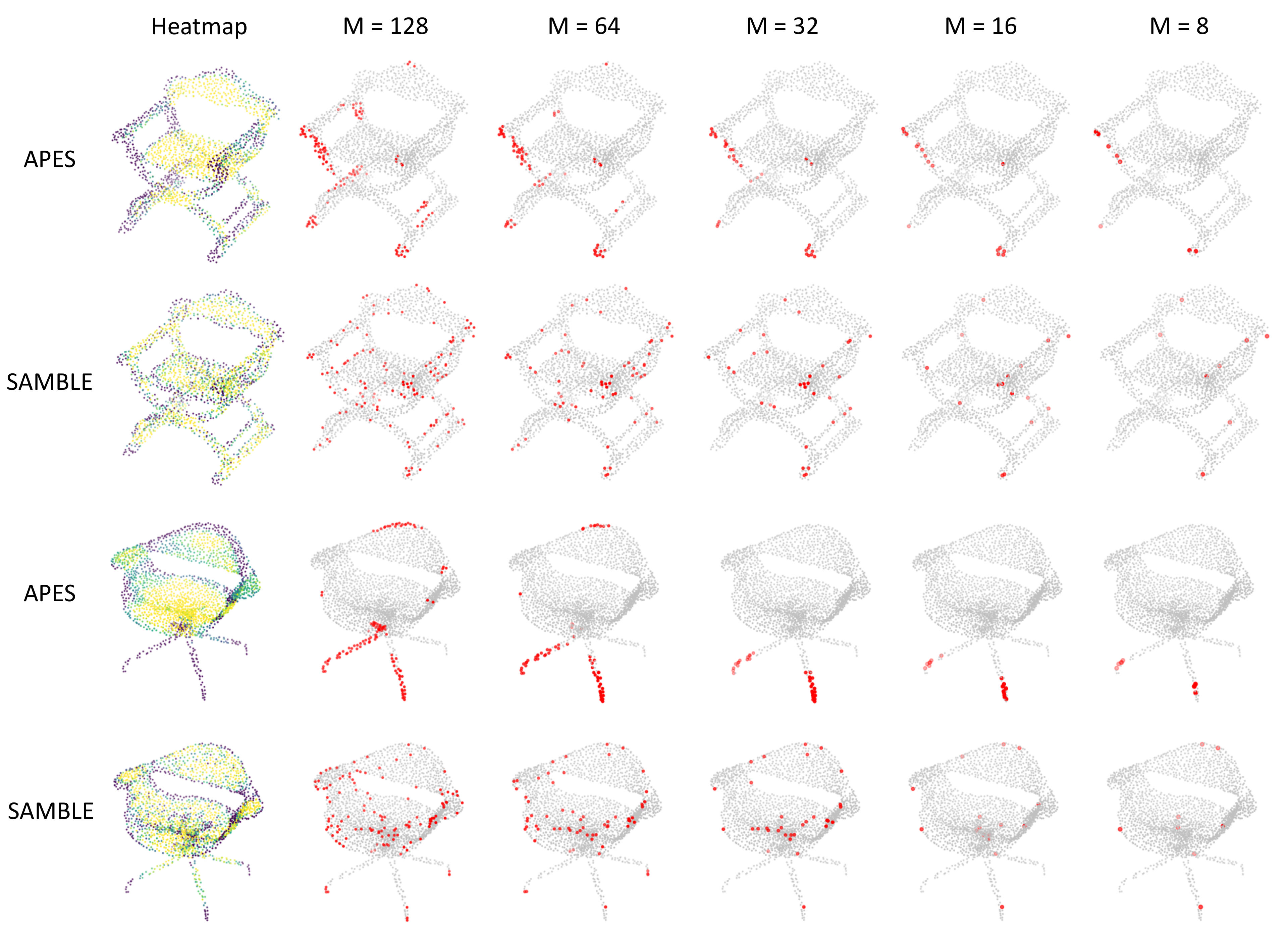}
    \caption{Sampled results of few-point sampling in comparison with APES. Zoom in for optimal clarity. \vspace{-0.1cm}}
    \label{fig:few}
\end{figure}

\vspace{2pt}
\noindent\textbf{Qualitative and Quantitative Results.}
Quantitative results are presented in \cref{tab:compare}. Note that APES \cite{wu2023attention} uses FPS to pre-process the input into $2M$ points while we do not. For a fair comparison, additional results of APES without the pre-processing step are also tested and reported. Nonetheless, even without pre-processing, SAMBLE achieves state-of-the-art results in the few-point sampling task as the number of sampled points decreases to extremely smaller ones.

Qualitative results are presented in \cref{fig:few}. For few-point sampling, APES relies on FPS to pre-sample the input into $2M$ points due to its limitations . In contrast, our method preserves better global uniformity, allowing direct few-point sampling from the input while still achieving satisfactory sampled results, as demonstrated in \cref{fig:few}. When sampling very few points, APES tends to concentrate on the sharpest regions, whereas our SAMBLE method preserves better global uniformity throughout the point cloud shape.

\subsection{Ablation Study}
\label{sec:ablation}
In this subsection, our emphasis is directed toward the novel designs introduced within this paper, excluding common topics such as network width. More ablation studies and further design justifications are provided in the supplementary materials to enhance our method's interpretability.

\vspace{2pt}
\noindent\textbf{Different Indexing Modes.}
Apart from the visualized heatmaps given in \cref{fig:indexingMode_carve}, we also report their respective experimental results in \cref{tab:indexing_mode}. The tests are performed using top-M as the sampling strategy. From it, we can observe that indexing modes vi and vii achieve best performances.

\begin{table}[t]
\centering
\resizebox{\linewidth}{!}{
\begin{tabular}{ccccccccc}
\toprule
\multicolumn{2}{c}{Indexing Mode}                  & i & ii & iii & iv & v  & vi & vii  \\ \midrule
Cls.                  & OA (\%)        & 93.92 & 93.78 & 93.63 & 93.66 & 93.40 & \textbf{94.11} & 94.08 \\ \midrule
\multirow{2}{*}{Seg.} &  Cat. mIoU (\%) & 83.98 & 83.85 & 83.62 & 83.51 & 83.47 & 84.12 & \textbf{84.22} \\

                      & Ins. mIoU (\%) & 86.16 & 85.99 & 85.74 & 85.60 & 85.49 & 86.38 & \textbf{86.46} \\ \bottomrule
\end{tabular}}
\caption{Classification and segmentation performance with different indexing modes.}
\label{tab:indexing_mode}
\end{table}

\begin{table}[t]
\centering
\resizebox{\linewidth}{!}{
\begin{tabular}{ccccccccc}
\toprule
\multicolumn{2}{c}{Number of Bins}           &  1 &  2 & 4 & 6 & 8 & 10 & 12    \\ \midrule
Cls.                  & OA (\%)  &  93.95 & 93.91 & 93.98 & \textbf{94.18} & 94.02 & 93.80 & 93.84   \\ \midrule
\multirow{2}{*}{Seg.} & Cat. mIoU (\%) & 84.22 & 84.14 & \textbf{84.51} & 84.40 & 84.19 & 83.98 & 84.36  \\
                      & Ins. mIoU (\%) & 86.46 & 86.28 & \textbf{86.67} & 86.61 & 86.48 & 86.23 & 86.43  \\ \bottomrule
\end{tabular}}
\caption{Classification and segmentation performance with different number of bins. \vspace{-0.2cm}}
\label{tab:study_bins}
\end{table}

\vspace{2pt}
\noindent\textbf{Number of Bins.} 
As a key parameter in SAMBLE, an ablation study is performed over the number of bins $n_b$. The results are presented in \cref{tab:study_bins}. Remarkably, increasing the number of bins does not yield improved performance. This phenomenon is likely attributable to the subdivision of shapes into an excessive number of point categories, leading to the gradual diminishment of score disparities across the bins. In our case, $n_b = 6$ and $4$ yield the best performance for the classification and segmentation tasks respectively, and we use it for the corresponding experiments.

\vspace{2pt}
\noindent\textbf{Upsampling layer.}
An important aspect to highlight is the upsampling layer. Most point cloud network models employ neighbor-based interpolation \cite{qi2017pointnet++,zhao2021point,qian2022pointnext} for upsampling, as FPS is used during the downsampling process. However, APES introduces a cross-attention layer for upsampling to address its limitations of overemphasizing edge points, which renders traditional neighbor-based interpolation impractical. In contrast, our method achieves an improved balance between sampling edge points and preserving global uniformity, allowing the use of interpolation operations during upsampling. An ablation study for evaluating various upsampling layers and interpolation with different $K_{up}$ values is conducted, and the results are presented in Table \ref{tab:upsample}. The results show a performance drop for APES when interpolation is used in place of cross-attention, while SAMBLE demonstrates superior performance with it.

\begin{table}[t]
\centering
\resizebox{\columnwidth}{!}{
\begin{tabular}{@{}ccccc@{}}
\toprule
\multirow{2}{*}{Upsample} & \multicolumn{3}{c}{Interpolation}                                                                 & \multirow{2}{*}{Cross-Attention} \\ \cmidrule(lr){2-4}
                          & \multicolumn{1}{c}{$K_{up} = 3$} & \multicolumn{1}{c}{$K_{up} = 8$} & \multicolumn{1}{c}{$K_{up} = 16$} &                                  \\ \midrule
APES (local)   & 82.89 / 85.40 & 82.95 / 85.44 & 82.96 / 85.42 & 83.11 / 85.58 \\
APES (global)  & 83.16 / 85.53 & 83.19 / 85.59 & 83.17 / 85.55 & 83.67 / 85.81 \\ \midrule\
SAMBLE         & \textbf{84.51 / 86.67} & 84.35 / 86.48 & 84.31 / 86.43 & 84.36 / 86.44 \\ \bottomrule
\end{tabular}
}
\caption{Segmentation results with different upsampling layers on ShapeNet-Part. The number before ``/'' is the category mIoU, and the number after is the instance mIoU. \vspace{-0.2cm}}
\label{tab:upsample}
\end{table}
\section{Conclusion}
\label{sec:conclusion}
In this paper, a novel point cloud sampling method SAMBLE is proposed to learn shape-specific sampling strategies for point cloud shapes. Based on a sparse attention map that integrates both local and global information, multiple indexing modes are designed and explored. By partitioning the points in each shape into bins and learning respective sampling ratios for each bin, shape-specific sampling strategies are acquired for individual point cloud shapes. SAMBLE achieves an optimal trade-off between sampling local details and preserving global uniformity, resulting in improved performance on downstream tasks. For future directions, advancements in upsampling layers could further improve the model's performance. Additionally, adapting the proposed method for point cloud scenes is another promising area to explore.
{
    \small
    \bibliographystyle{ieeenat_fullname}
    \bibliography{main}
}

\clearpage
\setcounter{page}{1}
\maketitlesupplementary

\section{Carve-based SAM and Insert-based SAM}
Based on different information basis, we propose two different sparse attention maps (SAM), carve-based SAM and insert-based SAM.
In the main paper, we used global information as the basis and introduced carve-based SAM. Using local information as the basis, insert-based SAM is introduced as follows.

\vspace{2pt}
\noindent\textbf{Insert-based Sparse Attention Map.} 
Carve-based sparse attention map starts from the global information, and then merges the local information. It can also be done in a reverse way: starting with the local information first, then considering it in a global situation. To be more specific, local-based attention maps are first computed with the equation presented in the main paper, subsequently, the values in the local attention map of each point are inserted in the corresponding cells of each row in an empty (initialized as all 0s) global $N \times N$ attention map based on the $k$-nearest neighbor indexes. We term it Insert-based Sparse Attention Map. Again, for each row, $k$ cells are inserted; and for each column, the number of inserted cells is not fixed.

\vspace{2pt}
\noindent\textbf{Relation between Carve and Insert-based SAM.} 
More vividly, consider the global attention map as a grid stone slab of size $N \times N$. For carve-based SAM, values of all cells are pre-computed and hidden in the slab grid cells, and only the selected cells are carved out; for insert-based SAM, only values of certain cells are pre-computed in the mosaic tile strings (the local-based attention maps), and they are then inserted into the slab according to the corresponding KNN indexes, like inserting mosaic tiles into an empty grid slate.
The final outputs from carve-based SAM and Insert-based SAM are quite similar since they have the same places of non-zero cells. For both methods, the number of selected cells in each row is always $k$, while the number of selected cells in each column is variable. Their main difference is that the row-wise sum in insert-based SAM is always 1, while in carve-based SAM is not.

\begin{figure}[t]
    \centering
    \includegraphics[width=1\linewidth,trim=18 0 20 0,clip]{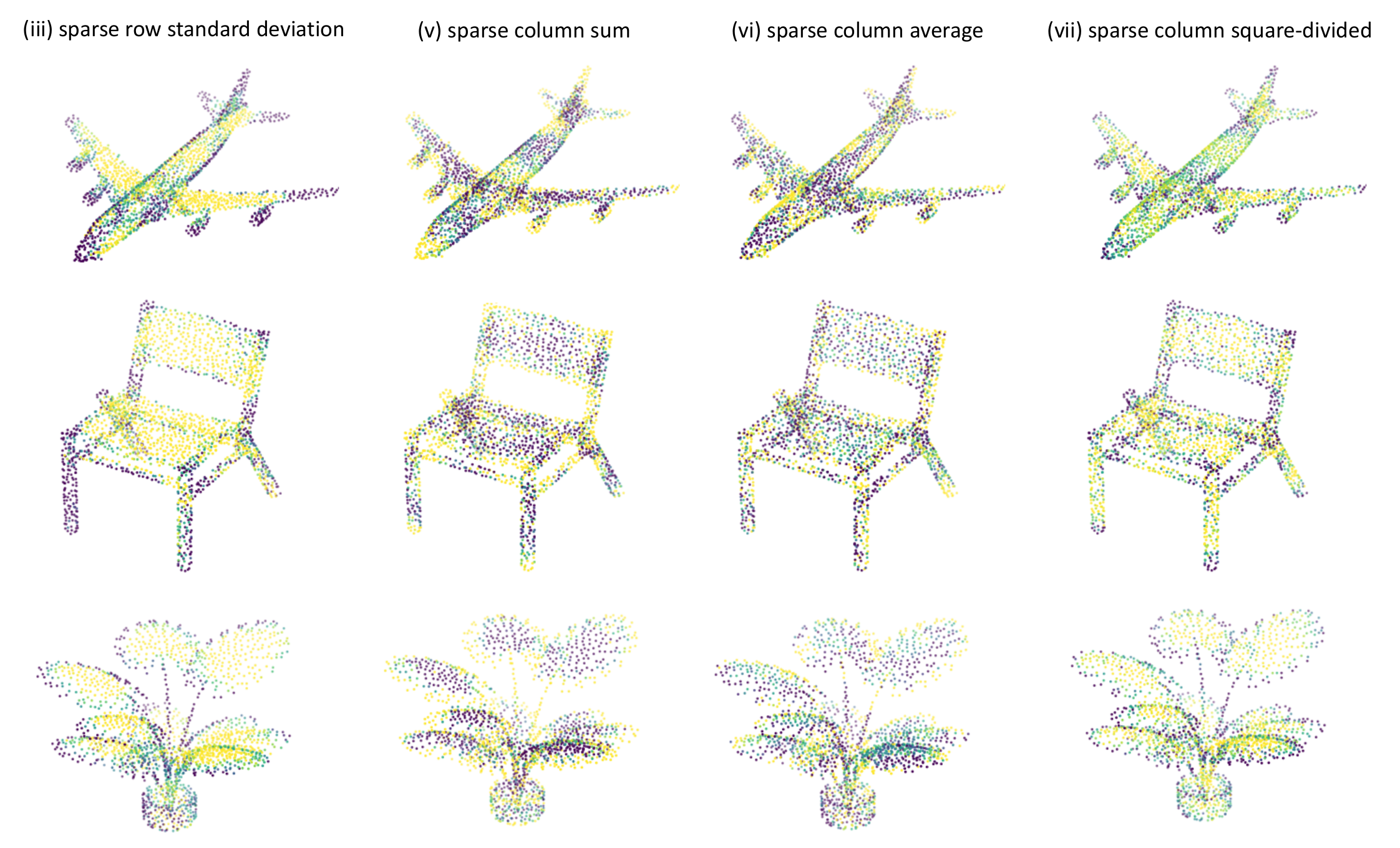}
    \caption{Heatmaps under different indexing modes with insert-based sparse attention map.}
    \label{fig:indexingMode_insert}
\end{figure}

\vspace{2pt}
\noindent\textbf{Different Indexing Modes with Insert-Based SAM.}  
Based on carve-based SAM, all seven indexing modes proposed are compatible. However, when insert-based SAM is used, since the sparse row-wise sum is always 1, only indexing modes iii, v,  vi, vii are compatible.

To investigate how each indexing mode works, we train a separate model for each indexing mode with all other settings consistent. In the paper, we have visualized the sampling score distributions as heatmaps for carve-based SAM to improve the method's interpretability. As supplementary materials, we also present such heatmap results with insert-based SAM as in \cref{fig:indexingMode_insert}. Note that only four indexing modes are applicable for insert-based SAM. The visualized results are quite similar to those of carve-based SAM except for indexing mode vi, with which insert-based SAM shows smaller differences between point sampling scores. On the other hand, indexing mode vii still achieves a better trade-off between sampling edge points and preserving global uniformity. However, the performance of insert-based SAM on downstream tasks is mostly not on par with carve-based SAM as presented in \cref{tab:samps_idxMode}, hence we use carve-based SAM as default for most experiments in our paper.

\begin{table}[t]
\centering
\resizebox{\linewidth}{!}{
\begin{tabular}{@{}ccccc@{}}
\toprule
\multirow{2}{*}{SAM} &
  \multirow{2}{*}{\begin{tabular}[c]{@{}c@{}} \rule{0pt}{2.5ex} Indexing\\ Mode\end{tabular}} &
  \multirow{2}{*}{\begin{tabular}[c]{@{}c@{}} \rule{0pt}{2.5ex} Cls.\\ OA (\%)\end{tabular}} &
  \multicolumn{2}{c}{Seg.} \\ \cmidrule(l){4-5} 
                        &     &                & Cat. mIoU (\%) & Ins. mIoU (\%) \\ \midrule
\multirow{7}{*}{Carve}  & i   & 93.92          & 83.98          & 86.16          \\
                        & ii  & 93.78          & 83.85          & 85.99          \\
                        & iii & 93.63          & 83.62          & 85.74          \\
                        & iv  & 93.66          & 83.51          & 85.60          \\
                        & v   & 93.40          & 83.47          & 85.49          \\
                        & vi  & \textbf{94.11} & 84.12          & 86.38          \\
                        & vii & 94.08          & \textbf{84.22} & \textbf{86.46} \\ \cmidrule(l){2-5} 
\multirow{4}{*}{Insert} & iii & 93.67          & 83.71          & 85.86          \\
                        & v   & 93.44          & 83.42          & 85.51          \\
                        & vi  & 93.46          & 83.64          & 85.78          \\
                        & vii & \textbf{93.83}          & \textbf{84.09}          & \textbf{86.15}          \\ \bottomrule
\end{tabular}
}
\caption{Classification and segmentation performance of different indexing modes with different SAMs. Top-M sampling is adopted as the sampling strategy.}
\label{tab:samps_idxMode}
\end{table}

\section{Determining Number of Sampled Points for Each Bin}
For each shape, by considering the number of points contained within bins $\boldsymbol{\beta} = (\beta_1, \beta_2, \ldots, \beta_{n_b})$ alongside the determined bin sampling weights $\boldsymbol{\omega} = (\omega_1, \omega_2, \ldots, \omega_{n_b})$, the specific numbers of points to be sampled from each bin $\boldsymbol{\kappa} = (\kappa_1, \kappa_2, \ldots,  \kappa_{n_b})$ are computed with the following algorithm. 

\vspace{0.2cm}
\begin{minipage}[b]{0.9\linewidth}
\hrule
\vspace{5pt}
\textbf{Algorithm 1} Determining $\boldsymbol{\kappa}$ from $\boldsymbol{\beta}$ and $\boldsymbol{\omega}$  
\vspace{5pt}
\hrule
\label{alg: determine k}
\begin{algorithmic}[1]
\Require {number of total points to be selected: $M$, Sampling weights 
$\boldsymbol{\omega}: [\omega_1, \omega_2, \ldots, \omega_{n_b}]$,
number of points in bins 
$\boldsymbol{\beta}: [\beta_1, \beta_2, ..., \beta_{n_b}] $} 
\State $\boldsymbol{\kappa} \gets \mathbf{0}$
\State $\bx \gets  \boldsymbol{\omega} \cdot \boldsymbol{\beta} + \epsilon$
\State $M_r \gets M$
\While{$ M_r > 0 $}
    \State $s \gets \frac{M_r}{\sum x_j}$
    \State \textbf{for} $j=1$ to $n_b$ \textbf{do}
    \State \hspace{\algorithmicindent} $\kappa_j \gets \mathrm{round}(\kappa_j + s x_j) $
    \State \hspace{\algorithmicindent} \textbf{if} $\kappa_j \geq \beta_j$ \textbf{then} 
    \State \hspace{\algorithmicindent} \hspace{\algorithmicindent} $\kappa_j \gets \beta_j$
    \State \hspace{\algorithmicindent} \hspace{\algorithmicindent} $x_j \gets 0$
    \State \hspace{\algorithmicindent} \textbf{end if}
    \State \textbf{end for}
    \State $M_r \gets M - \sum \kappa_j$
\EndWhile
\State \Return $\boldsymbol{\kappa}$
\end{algorithmic}
\hrule
\end{minipage}
\vspace{0.2cm}

\begin{figure}[h]
    \centering
    \includegraphics[width=1\linewidth,trim=2 2 2 2,clip]{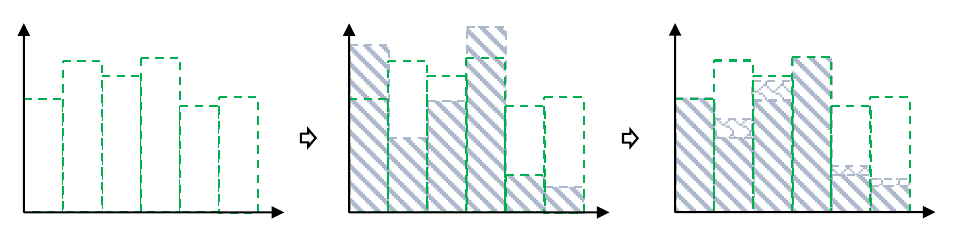}
    \caption{Illustration of redistributing excess points to other bins that have not been fully selected.}
    \label{fig:redistribute}
\end{figure}

In the above algorithm, $s$ in line 5 is the scaling factor used to ensure a desired number of total sampled points. It is also evident that our sampling method is scalable to any desired sampling ratio. In line 2, $\epsilon$ is a minimal value (here we use $1 \times 10^{-8}$) to prevent the denominator part from being zero in a later step.
Moreover, to prevent $\kappa_j$ from surpassing the available number $\beta_j$ in any bin, any excesse points are proportionately redistributed to other bins that have not been fully selected. The redistribution process is further illustrated in \cref{fig:redistribute} for better comprehensibility.

\section{Relationship between Bin Sampling Weights and Bin Sampling Ratios}
\begin{figure}[h]
    \centering
    \includegraphics[width=0.9\linewidth,trim=0 5 0 5,clip]{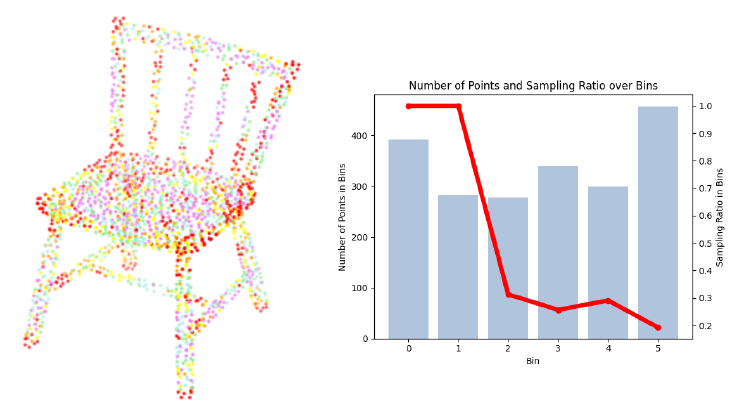}
    \caption{Left: bin partitioning, each color represents the points belonging to this bin. Right: the learned sampling strategy.}
    \label{fig:weightToRatio}
\end{figure}

For the sake of brevity and improved visual clarity, in the paper, the axis labels of the histograms have been omitted. We further provide the full version of the histogram, in which the number of points and the sampling ratio in each bin are given. A demo is provided in \cref{fig:weightToRatio}. More detailed histogram results are provided in Sec. \ref{sec:moreHistVis}.

One thing worth noting is that the indicated sampling ratios $\br$ in the histogram are not simply re-scaled sampling weights $\boldsymbol{\omega}$. As in the algorithm we presented before, apart from the re-scaling operation, a redistribution operation is also applied to prevent $\kappa_j$ surpassing the available point number $\beta_j$ in one bin. 
Given the point number in each bin $\boldsymbol{\beta} = (\beta_1, \beta_2, \ldots, \beta_{n_b})$ and the number of points to be sampled from each bin $\boldsymbol{\kappa} = (\kappa_1, \kappa_2, \ldots,  \kappa_{n_b})$, the sampling ratios presented in the histogram is $\br = \boldsymbol{\kappa} / \boldsymbol{\beta}$ and $\br \in [0, 1]$.

\begin{table}[h]
\centering
\resizebox{1\linewidth}{!}{
\begin{tabular}{ccccccc}
\toprule
Bin Index & 0 & 1 & 2 & 3 & 4 & 5    \\ \midrule
\begin{tabular}[c]{@{}l@{}} Possibilities of All \\ Points Being Sampled \end{tabular} & 53.69\% & 27.11\% & 8.02\% & 2.11\% & 0.85\% & 4.98\% \\  \bottomrule
\end{tabular}}
\caption{Possibilities of all points being sampled in bins, across all shapes from the test dataset.}
\label{tab:binSurpass}
\end{table}

The redistribution operation only happens when $\kappa_j$ is about to surpass $\beta_j$, this means all points in $j$th bin have been selected and $r_j = 1$. We additionally count and document the likelihood of this occurrence for all bins across all shapes from the test dataset. The numbers are reported in \cref{tab:binSurpass}, from which we can find that for around 54\% of the shapes, all points in the first bin are selected and sampled. Note that the first bin corresponds to the points of higher sampling scores which are mostly edge points with indexing mode vii. This observation underscores the significance of edge points. On the other hand, there are still around 46\% shapes that do not sample all edge points. It suggests that an excessive emphasis on edge points might have adverse effects on subsequent downstream tasks for these shapes, which also aligns with the conclusion drawn by APES.

\begin{figure*}[t]
    \centering
    \includegraphics[width=1\linewidth,trim=0 0 0 0,clip]{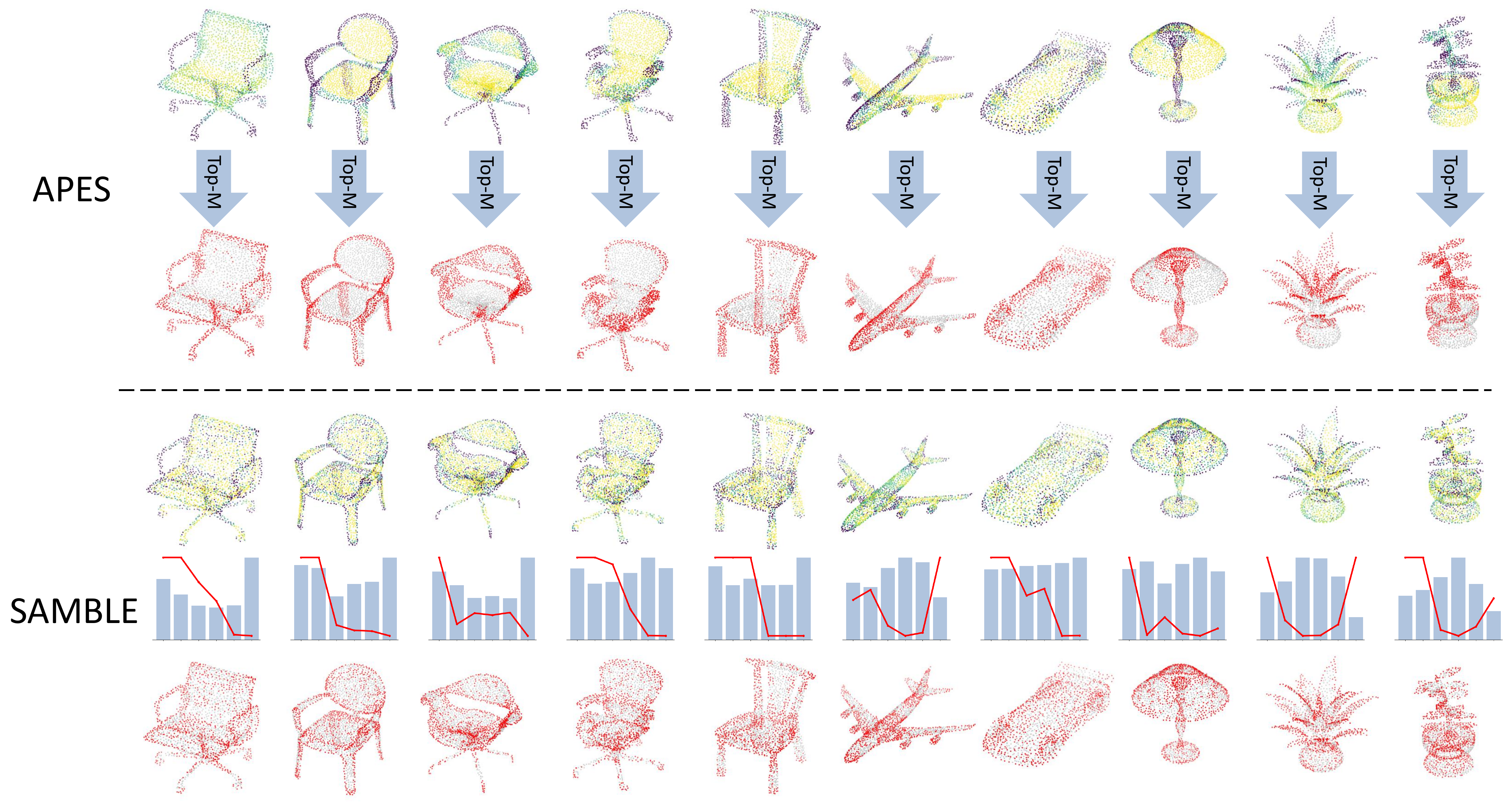}
    \caption{Qualitative results of our proposed SAMBLE, in comparison with APES. In addition to the sampled results, sampling score heatmaps and sampling strategies are also provided. All shapes are from the test set.}
    \label{fig:cls_supp}
\end{figure*}

\section{Network Architecture}
\begin{figure}[h]
    \centering
    \includegraphics[width=1\linewidth,trim=2 2 2 2,clip]{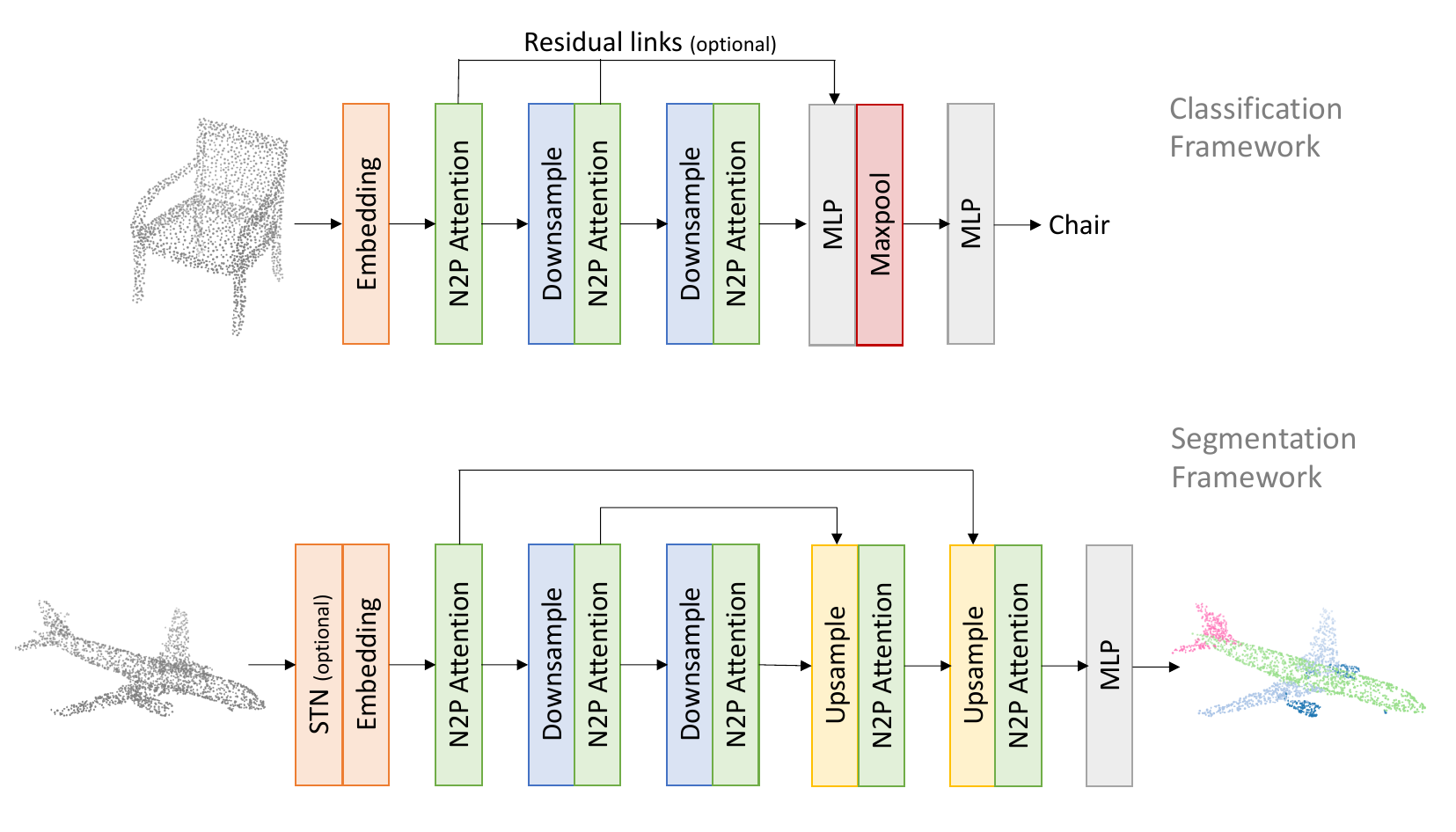}
    \caption{Network architectures for the classification task and the segmentation task.}
    \label{fig:net}
\end{figure}

For a fair comparison, the same basic network architectures from APES are used in our experiments, as illustrated in \cref{fig:net}. The downsampling layers are replaced with our proposed ones, and the upsampling layers are replaced with the classical interpolation-based ones.

\section{More Training Details}

\textbf{Classification Tasks.}
AdamW is used as the optimizer. The learning rate starts from $1\times 10^{-4}$ and decays to $1\times 10^{-8}$ with a cosine annealing schedule. The weight decay hyperparameter for network weights is set as $1$. Dropout with a probability of $0.5$ is used in the last two fully connected layers. We use $n_b = 6$ bins for point partitioning. The momentum update factor $\gamma = 0.99$ for updating boundary values. The temperature parameter $\tau = 0.1$. The network is trained with a batch size of 8 for 200 epochs.

\vspace{2pt}
\noindent\textbf{Segmentation Tasks.}
AdamW is used as the optimizer. The learning rate starts from $1\times 10^{-4}$ and decays to $1\times 10^{-8}$ with a cosine annealing schedule. The weight decay hyperparameter for network weights is $1\times 10^{-4}$. We use $n_b = 4$ bins for point partitioning. The momentum update factor $\gamma = 0.99$ for updating boundary values. The temperature parameter $\tau = 0.1$. The network is trained with a batch size of 16 for 200 epochs.

\section{Sampling Results in Comparison with APES}
\begin{figure*}[t]
    \centering
    \includegraphics[width=1\linewidth,trim=0 0 0 0,clip]{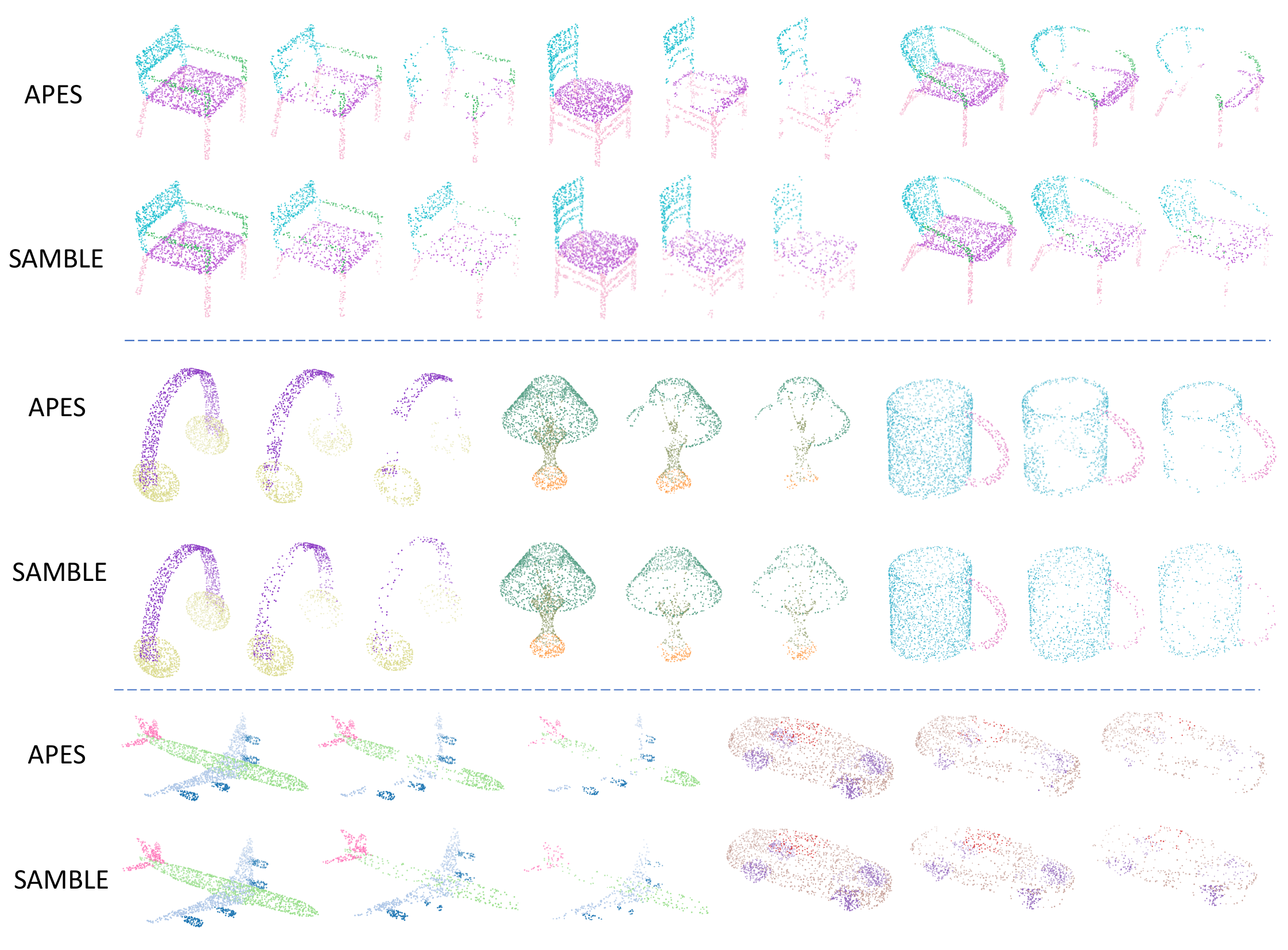}
    \caption{Segmentation results of our proposed SAMBLE, in comparison with APES. All shapes are from the test set.}
    \label{fig:seg_supp}
\end{figure*}

Additional qualitative results in comparison with APES are provided in \cref{fig:cls_supp} and \cref{fig:seg_supp}. Both figures indicate that APES focuses excessively on edge points, while SAMBLE successfully achieves a much better trade-off between sampling edge points and preserving global uniformity, leading to better performance on downstream tasks.

\section{Design Justifications of the Bin Token Idea - Devil Is in the Details.}
\textbf{Adding Bin Tokens to Q or K/V?}
A critical point in the idea of bin tokens lies in determining the specific branches to which the tokens should be concatenated. In order to match the tensor dimension for later computation in the attention mechanism, the tensor size of Key and Value should be the same. Hence if tokens are being added to the Key branch, they also have to be added to the Value branch. Overall, there are two possibilities of adding bin tokens to (i) the Query branch, or (ii) the Key and the Value branches.

It is crucial to emphasize that, due to the nature of the sampling operation where indexes are selected, gradients cannot be propagated back through the sampling operation during the backward propagation process. As a result, regardless of the selected structure, it is essential to establish an alternative pathway to convey the information contained within the bin tokens, which have a size of $n_b \times N$, to the downsampled features, which have a size of $M \times d$. This pathway should ensure the flow of relevant information despite the inability to directly backpropagate gradients through the sampling operation.

\begin{figure}[h]
    \centering
    \includegraphics[width=\linewidth,trim=0 0 0 0,clip]{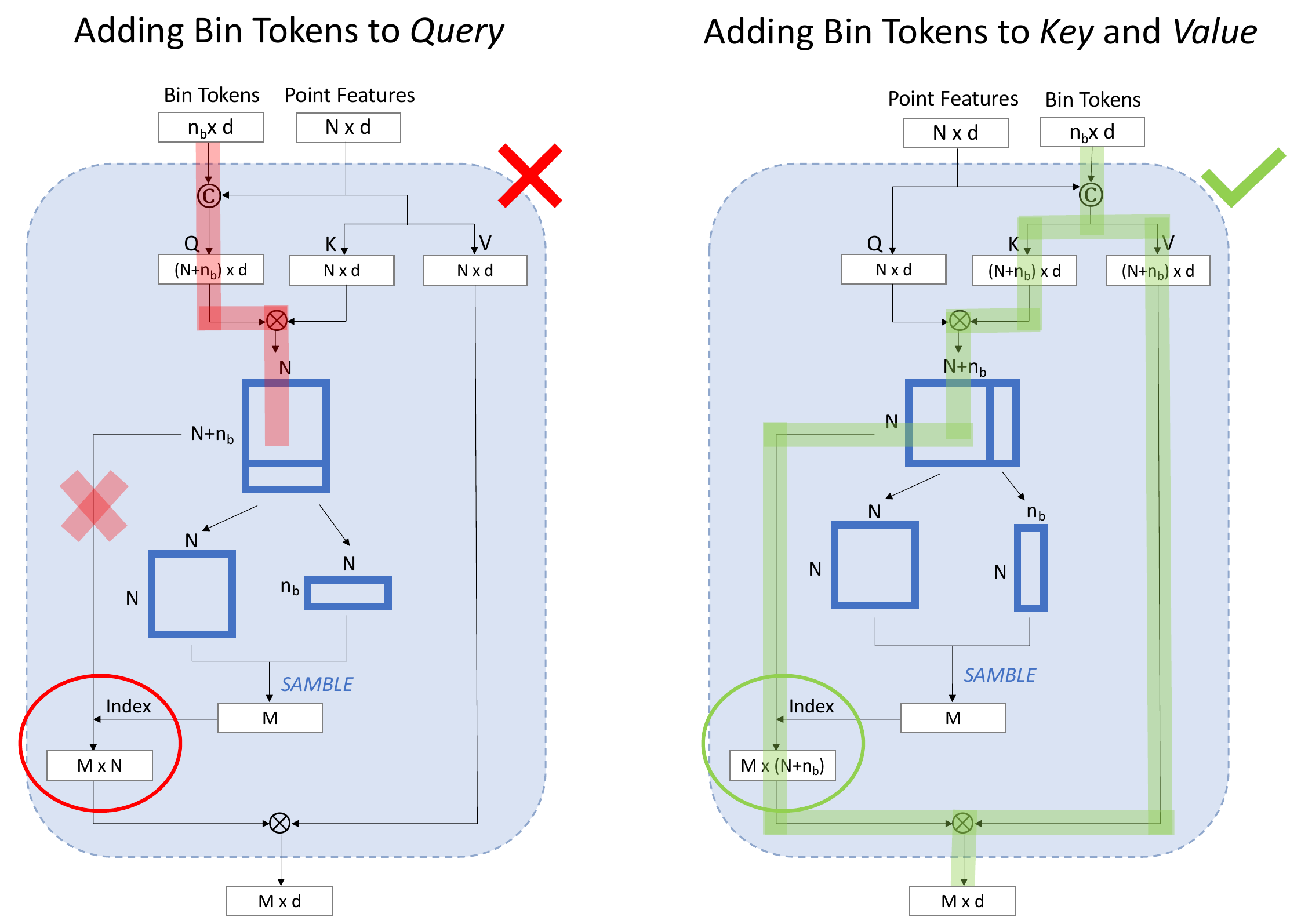}
    \caption{Adding bin tokens to Query leads to no gradient being backpropagated to the tokens, while adding bin tokens to Key and Value enables the gradient backpropagation.}
    \label{fig:6NorN6}
\end{figure}

As illustrated in the left of \cref{fig:6NorN6}, in the former case, an attention map of tensor size $(N + n_b) \times N$ is obtained. After $M$ indexes of the points to be sampled are learned with SAMBLE, $M$ rows in the attention map are extracted to form a new tensor for the next steps. However, note that the sub-tensor of $n_b \times N$ will never be delivered to the next steps since they do not correspond to points, hence no gradient will be backpropagated to the tokens during the training.

On the other hand, as illustrated in the right of \cref{fig:6NorN6}, adding bin tokens to the Key and Value branches does not have this problem and successfully enables gradient backpropagation. One thing worth mentioning is that in this scenario, the row-wise sum is not exactly equal to 1 but still very close to 1 due to the significantly smaller magnitude of $n_b$ relative to $N$. Therefore, this is unlikely to significantly impact the calculation of point-wise sampling scores. Concerning the design of adding bin tokens to all branches of Query, Key, and Value, it is equivalent to case ii since the sub-tensor of $n_b$ rows in the attention map will never be sampled and propagated.

\vspace{2pt}
\noindent\textbf{Order of Mean-pooling and ReLU Operations.}
Within our design, the ReLU operation is used to prevent the learned sampling weight from being negative. It can be performed either after Mean-pooling (as done in the main paper), or before Mean-pooling:
\begin{equation}
\label{equ:relu_mean}
    \omega_{j} = \frac{1}{\beta_j}\sum_{\bp_i \in \cB_j} \text{ReLU}(m_{\bp_i,\cB_j}) \, .
\end{equation}
However, the inherent distribution of values within tensors often results in a non-negligible proportion being negative, especially those corresponding to points of lower importance. Directly setting too many values to zero would result in a significant loss of features, which is regrettable considering the potential information discarded. Therefore, instead of performing the ReLU operation before the mean-pooling operation, we do it the other way around, i.e., first mean-pooling, then, after this information fusion, ReLU is performed over the pooled results.

\begin{figure*}[t]
    \centering
    \includegraphics[width=0.9\linewidth,trim=0 0 0 0,clip]{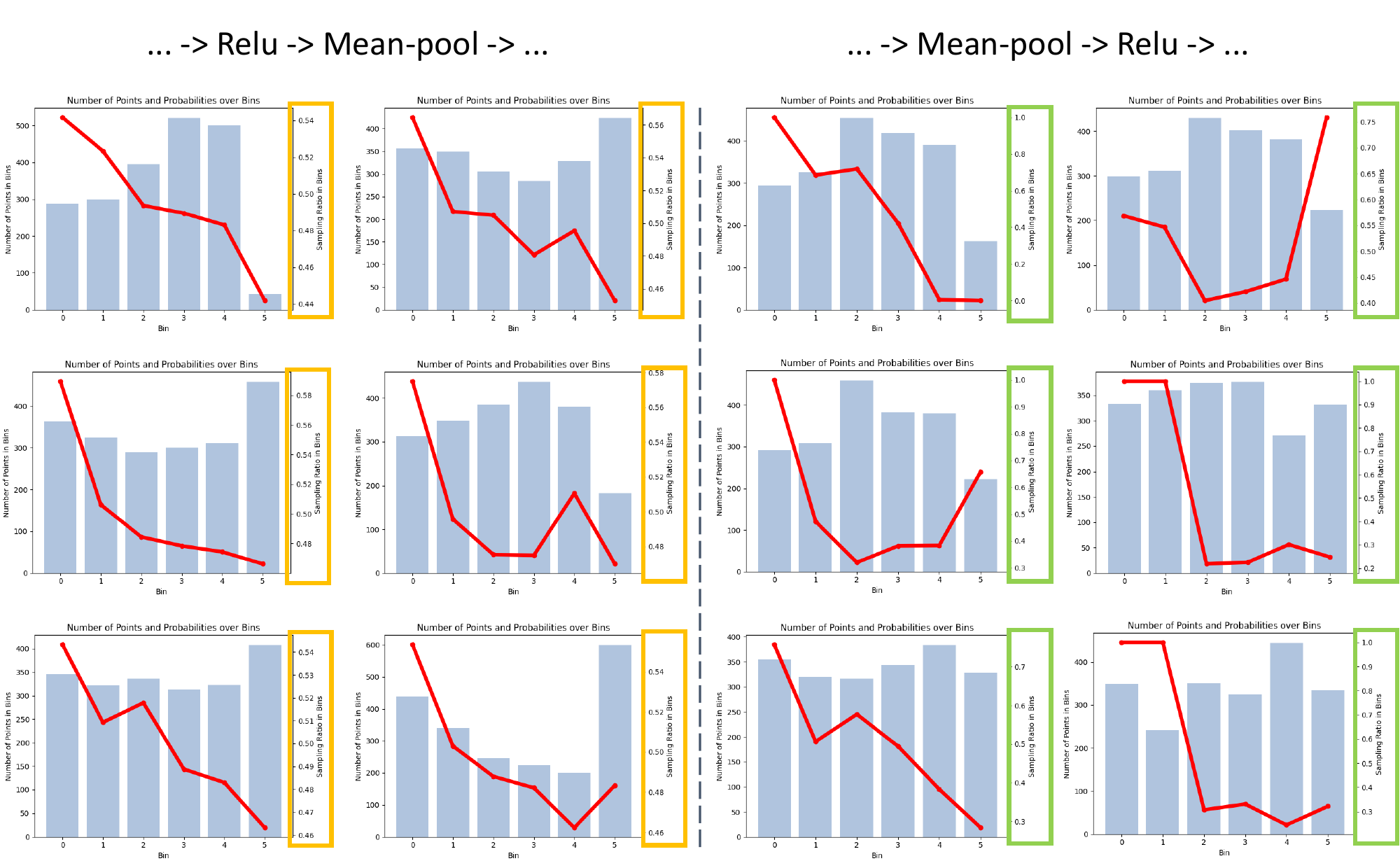}
    \caption{Learned sampling strategies with the mean-pooling and ReLU operations applied in different orders.}
    \label{fig:meanReLU}
\end{figure*}

\cref{fig:meanReLU} gives the learned sampling strategies with the mean-pooling and ReLU operations applied in different orders. Although both orders yield shape-specific sampling strategies, the sampling ratios over bins learned with the order of ReLU first are mostly around 40\% - 60\%, leading to a worse sampling performance. On the other hand, the order of mean-pool first yields better sampling strategies as less potential information is discarded.

We additionally count and document the likelihood of ReLU being effective, which indicates the former pooled result is negative, for all bins across all test shapes. From the numbers reported in \cref{tab:binNegative}, we can see that the likelihood of the pooled results being negative is extremely small (less than 1\%) for the first half of bins, while it goes higher for the latter bins yet the number is still relatively acceptable. 

\begin{table}[h]
\centering
\resizebox{1\linewidth}{!}{
\begin{tabular}{ccccccc}
\toprule
Bin Index & 0 & 1 & 2 & 3 & 4 & 5    \\ \midrule
\begin{tabular}[c]{@{}l@{}} Possibilities of \\ ReLU Being effective \end{tabular} & 0.45\% & 0.28\% & 0.57\% & 4.25\% & 11.63\% & 13.53\% \\  \bottomrule
\end{tabular}}
\caption{Possibilities of ReLU being effective in bins, across all test shapes.}
\label{tab:binNegative}
\end{table}

\vspace{2pt}
\noindent\textbf{Pre-softmax or Post-softmax Attention Map for Splitting The Point-to-Token Sub-Attention Map.}
When addressing the bin tokens, our initial approach involved splitting the point-to-token sub-attention map from the post-softmax attention map $\bM_{\text{post}}$, which seemed intuitively appropriate. Furthermore, all elements within $\bM_{\text{post}}$ are inherently positive, eliminating any concern for negative sampling weights and obviating the need for an additional ReLU operation. However, experimental findings revealed that this method proved ineffective, as it resulted in overly uniform sampling weights across different bins.

The underlying cause of this issue was identified after we explored the underlying mathematical principles and examined the values in the tensors during runtime. Tensors in a well-trained network tend to exhibit diminutive feature values as they propagate through layers. Denote $m_{ij}$ as one element in the pre-softmax attention map $\bM_{\text{pre}}$, given its minute magnitude, we apply the Taylor expansion formula to yield:
\begin{equation}
\label{equ:Taylor}
e^{m_{ij}} = 1 + m_{ij} + \frac{m_{ij}^2}{2} + \cdots  \approx 1 + m_{ij} \, .
\end{equation}
Therefore, the corresponding element $m'_{ij}$ in the post-softmax attention map is
\begin{equation}
\label{equ:Taylor_for_softmax}
m'_{ij} = \frac{e^{m_{ij}}}{\sum_{j=1}^{N + n_b}e^{m_{ij}}} \approx \frac{1 + m_{ij}}{N + n_b + \sum_{j=1}^{N + n_b}{m_{ij}}} \, .
\end{equation}

In our case, the values of the elements $m_{ij}$ in $\bM_{\text{pre}}$ are approximately within the magnitude of $10^{-3}$ to $10^{-5}$. After a softmax operation, the resultant values $m'_{ij}$ in $\bM_{\text{post}}$ exhibit minimal variation, leading to closely similar sampling weights across bins in a later step.

\begin{figure}[t]
    \centering
    \includegraphics[width=0.8\linewidth,trim=0 0 0 0,clip]{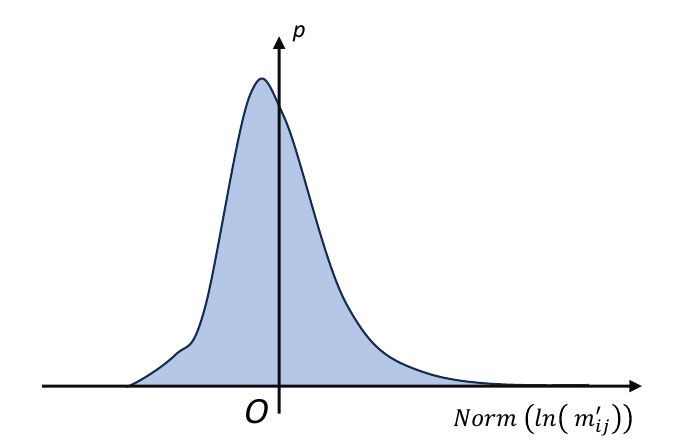}
    \caption{Illustrative figure of the distribution of the element values in the post-softmax attention map, after normalization.}
    \label{fig:distribution}
\end{figure}

Efforts were undertaken to address this issue before we turned to using $\bM_{\text{pre}}$ for sampling weights acquisition. We attempted to use the logarithmic operation to restore the lost information:
\begin{equation}
\label{equ:log_for_softmax}
\text{ln}(m'_{ij}) = \text{ln}(\frac{e^{m_{ij}}}{\sum_{j=1}^{N + n_b}e^{m_{ij}}}) = m_{ij} - \text{ln}(\sum_{j=1}^{N + n_b}e^{m_{ij}})
\end{equation}
After the logarithmic operation, every value in the sub-attention map is negative. Therefore, a normalization operation is necessary. However, as shown in \cref{fig:distribution}, the common normalization methods, such as z-score and centering, will result in too many negative elements (more than half), leading to too much information loss when passing through subsequent ReLU modules. Even if we successfully identify or meticulously design a superior normalization method that enables manual control over the proportion of negative elements to an applicable value, such manual intervention strays from the original intention of this thesis, which is to discover a learning-based mapping from sampling score to sampling probability.

Through the analysis, we observed that the term $m_{ij}$ in \cref{equ:log_for_softmax} is exactly the elements in the pre-softmax attention map and is what we are interested in. Therefore, to avoid the potential loss of information that could arise from the softmax operation, we opted to directly use the results from $\bM_{\text{pre}}$ for bin sampling weights acquisition.

\section{Additional Ablation Studies}
\textbf{Momentum Update Factor.} 
The momentum update strategy is widely used within contrastive learning frameworks in self-supervised learning. In our case, we aim to derive the bin boundary values $\boldsymbol{\nu}$ from the entirety of shapes within the training dataset. These values aim to evenly partition the distribution of point sampling scores across all shapes and points in the training data. Hence such an adaptive learning method is used.

An ablation study over the momentum update parameter $\gamma$ is performed and the numerical results are reported in \cref{tab:gamma}. From it, we can see that $\gamma = 0.99$ yields the best performance. This actually aligns with most current contrastive learning frameworks, where a majority use a value of $\gamma = 0.99$.

\begin{table}[t]
\centering
\resizebox{0.8\linewidth}{!}{
\begin{tabular}{cccccc}
\toprule
\multicolumn{2}{c}{$\gamma$}                  & 0.9  & 0.99  & 0.999  & 0.9999     \\ \midrule
\, Cls. & OA (\%) & \, 93.80 \, & \, \textbf{94.18} \, & \, 94.02 \, & \, 93.95 \, \\  \bottomrule
\end{tabular}}
\caption{Classification performance with different values of the momentum update factor $\gamma$.}
\label{tab:gamma}
\end{table}

\begin{figure}[t]
    \centering
    \includegraphics[width=0.8\linewidth,trim=0 0 0 0,clip]{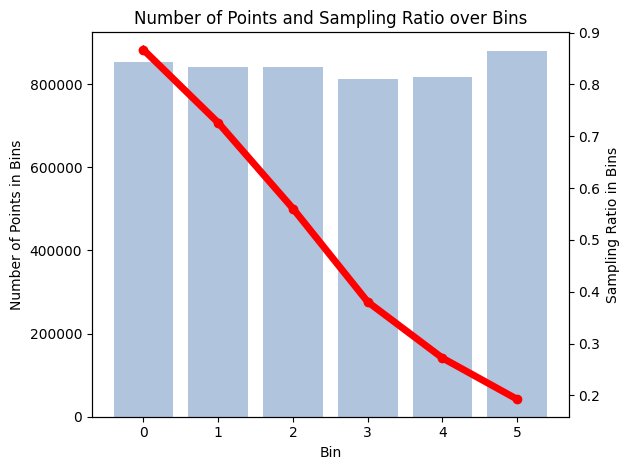}
    \caption{Partitioning the distribution of point sampling scores of all shapes and points in the test dataset into bins with the learned boundary values.}
    \label{fig:hisAll}
\end{figure}

We additionally provide the bin partitioning results over the test dataset with the learned boundary values $\boldsymbol{\nu}$ in \cref{fig:hisAll}. It demonstrates that the boundary values adaptively learned from the training dataset can also effectively partition the distribution of point sampling scores evenly across all shapes and points in the test dataset.

\vspace{2pt}
\noindent\textbf{Temperature Parameter.} 
The sampling strategy is determined with the point number in each bin $\boldsymbol{\beta} = (\beta_1, \beta_2, \ldots, \beta_{n_b})$ and the number of points to be sampled from each bin $\boldsymbol{\kappa} = (\kappa_1, \kappa_2, \ldots,  \kappa_{n_b})$.
Within each bin, instead of applying the top-M sampling method simply, we suggest employing random sampling with priors. The idea is quite straightforward: process the point-wise sampling scores into point-wise sampling probabilities, and $M$ non-repeated points are sampled randomly based on their sampling probabilities: 
\begin{equation}
\label{equ: tau}
\rho_{\bp_i} = \frac{e^{{a_{\bp_i}}/\tau}}{\sum_{i=1}^N e^{{a_{\bp_i}}/\tau}} \, ,
\end{equation}
where the temperature parameter $\tau$ controls the distribution of the sampling probabilities. 

\begin{figure}[h]
    \centering
    \includegraphics[width=1\linewidth,trim=0 0 0 0,clip]{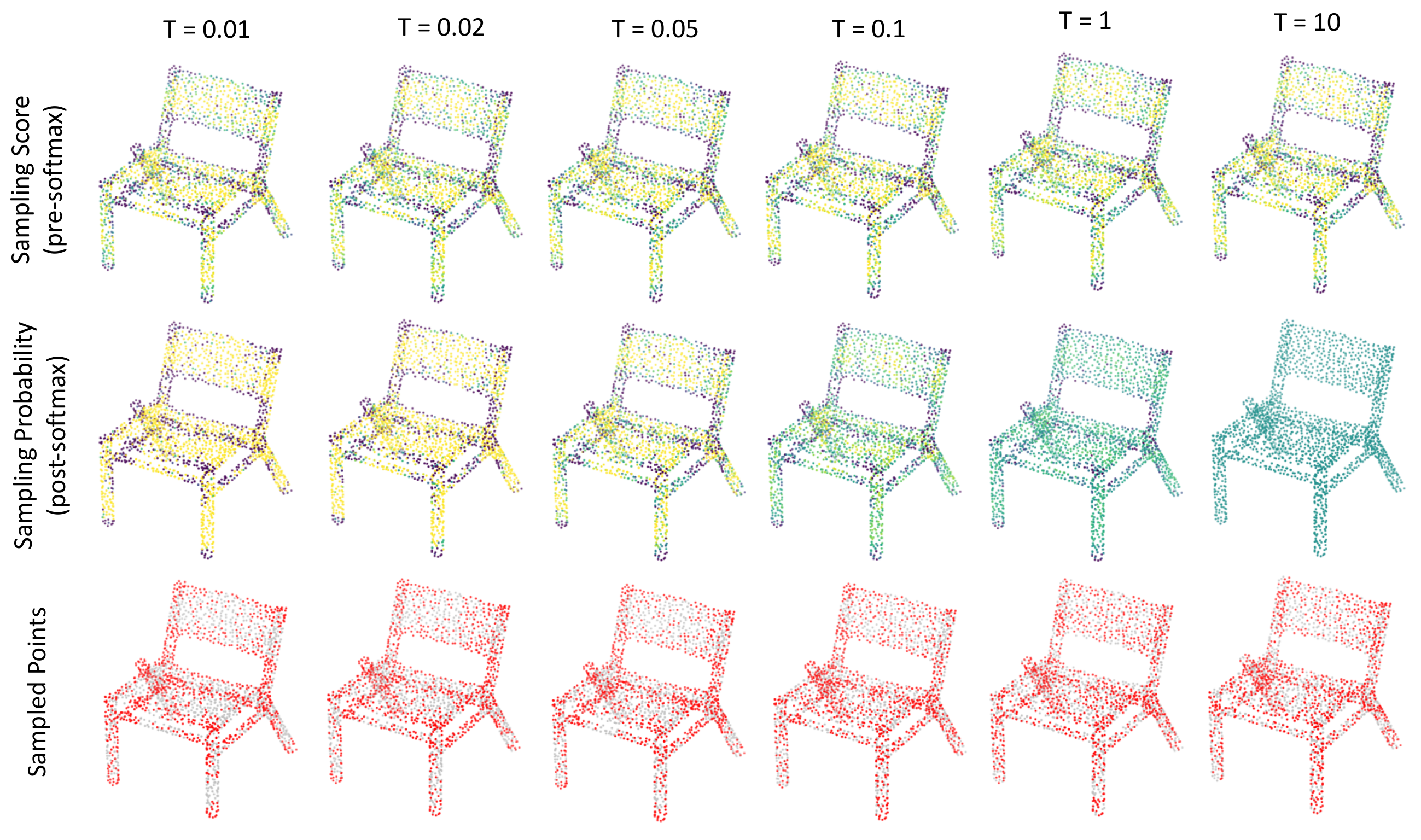}
    \caption{Different sampling results using different $\tau$ in the softmax with temperature during the sampling process. The indexing mode is the sparse column square-divided.}
    \label{fig:boltzmann}
\end{figure}

An ablation study over $\tau$ has been conducted. For a better illustration, the pre-softmax point sampling score heatmap and the post-softmax point sampling probability heatmap are visualized in \cref{fig:boltzmann}. However, since the softmax operation is performed within each bin, it would be impossible to visualize the post-softmax sampling probabilities of different bins in the same figure if multi-bins are used. Hence in \cref{fig:boltzmann} only a single bin is used. From it, we can observe that the sampling probabilities of points go from having a large deviation to being uniformly distributed, just as we designed. Numerical results are reported in \cref{tab:tau}, where $\tau=0.1$ achieves the best performance.

\begin{table}[t]
\centering
\resizebox{1\linewidth}{!}{
\begin{tabular}{cccccccccc}
\toprule
\multicolumn{2}{c}{$\tau$}             & 0.01  & 0.02  & 0.05  & 0.1 & 0.2 & 0.5  & 1     & 10    \\ \midrule
Cls.                  & OA (\%)        & 93.84  & 93.96 & 94.06 & \textbf{94.18} & 93.89 & 93.84 & 93.74 & 93.70 \\ \midrule
\multirow{2}{*}{Seg.} & Cat. mIoU (\%) & 84.10 & 84.23 & 84.38 & \textbf{84.51} & 84.26 & 84.13 & 84.02 & 83.88 \\
                      & Ins. mIoU (\%) & 86.44 & 86.48 & 86.60 & \textbf{86.67} & 86.51 & 86.42 & 86.29 & 86.23 \\ \bottomrule
\end{tabular}}
\caption{Classification and segmentation performance of the model with different $\tau$ values.}
\label{tab:tau}
\end{table}

\section{Sampling Policy Comparison}
\begin{figure}[t]
    \centering
    \includegraphics[width=1\linewidth,trim=0 0 0 0,clip]{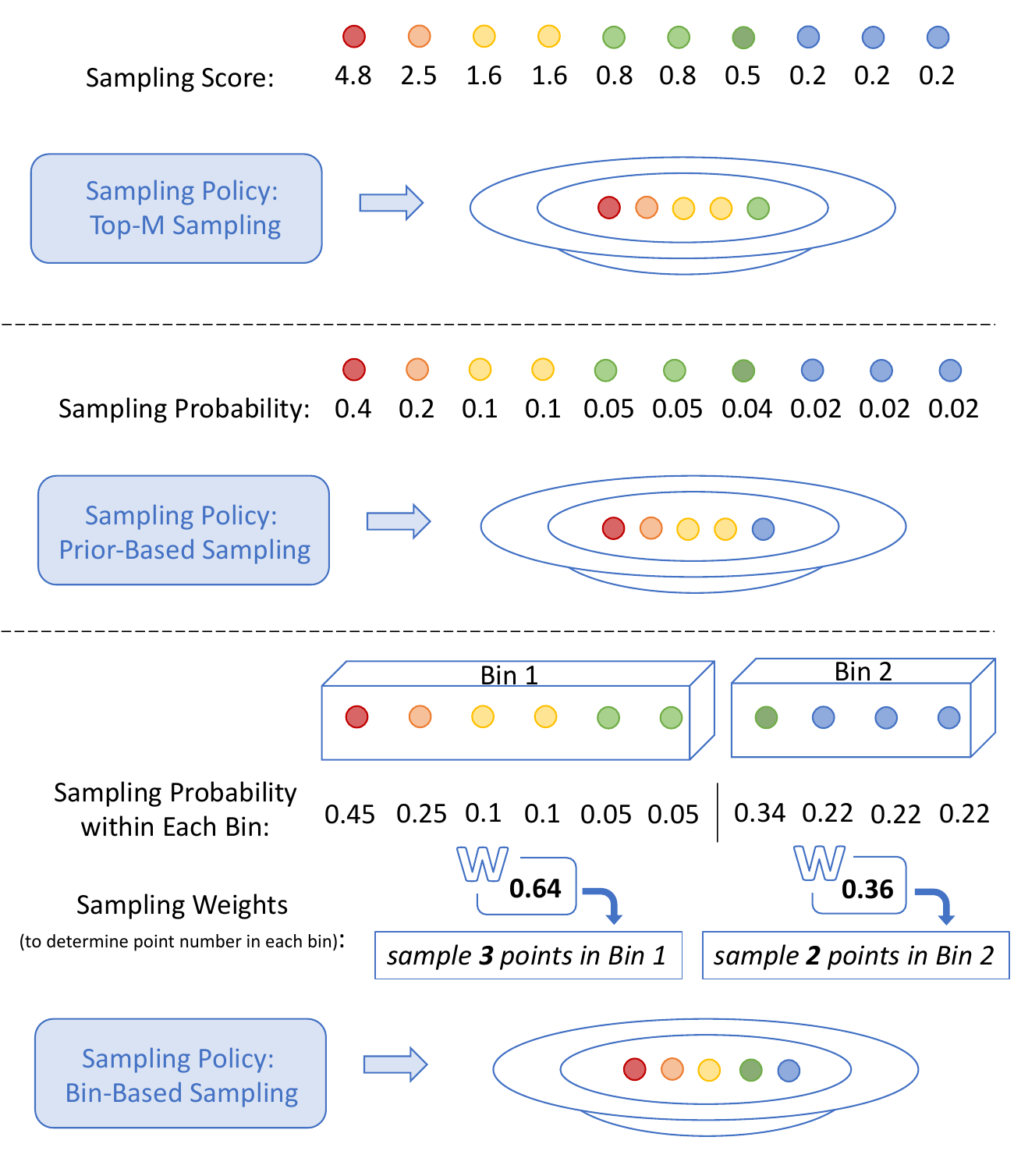}
    \caption{An illustration of different sampling policies. Note for bin-based sampling, either top-M sampling or prior-based sampling may be used within each bin. \vspace{0pt}}
    \label{fig:concept_policy}
\end{figure}

Three different sampling policies are illustrated in \cref{fig:concept_policy}, including Top-M sampling, prior-based sampling, and bin-based sampling. The Top-M sampling policy is the simplest one and it samples the points with larger sampling scores directly. The prior-based sampling policy first converts the sampling scores into sampling probabilities and then samples randomly based on those probabilities. This method introduces the possibility of allowing the sampling of points with smaller sampling scores. The bin-based sampling policy further builds upon that. It first partitions the points into bins, and then learns bin sampling weights to determine the number of points to be sampled within each bin, In this way, it guarantees the sampling of some smaller-score points if the model thinks they are helpful for downstream tasks. In each bin, either top-M sampling or prior-based sampling can employed. In our case, we use the prior-based sampling. The bin-based sampling policy allows for more fine-grained control over the sampling process, tailoring it to the specific characteristics of each shape.

\section{Model Complexity and Runtime Efficiency}
To evaluate SAMBLE's efficiency, we assess its model complexity in comparison with APES and report the results in \cref{tab:samble_modelComplexity}. Results from the traditional FPS and SAMBLE's variations are also reported. For a more direct and detailed comparison, we report both the number of parameters and FLOPS of a single downsampling layer. In order to assess inference efficiency, experiments were carried out using a trained ModelNet40 classification model on a single NVIDIA GeForce RTX 3090. The tests were conducted with a batch size of 8, evaluating a total number of 2468 shapes from the test set.

As shown in \cref{tab:samble_modelComplexity}, SAMBLE has a slightly larger number of model parameters compared to APES, primarily due to the incorporation of additional bin tokens. Notably, when $n_b=1$, the number of parameters and FLOPs of SAMBLE are identical to that of global-based APES. This is quite reasonable as in this case, using additional bin tokens is unnecessary and the multi-bin-based sampling policy degrades into the simple prior-based sampling policy (see \cref{fig:concept_policy}). On the other hand, SAMBLE's inference throughput is reduced due to the introduction of bin partitioning operations. Notably, the process of determining the number of points to be sampled within each bin involves a CPU-intensive loop computation (the redistribution process in Algorithm 1), which can lead to increased inference time. 

Overall, using a sparse attention map instead of a full attention map improves performance on downstream tasks, though it slightly decreases inference throughput. Similarly, replacing top-M sampling with prior-based sampling for the sampling policy shows comparable results—enhancing performance at the cost of a minor reduction in efficiency. In contrast, when bin-based sampling is employed, there is a significant boost in model performance, but this comes with a notable decrease in efficiency. Despite this, SAMBLE consistently outperforms FPS in both performance and speed. Note that the FPS results presented here already use a GPU-accelerated version, whereas its standard implementation achieves a much lower throughput (around 12).

Considering the trade-off between model performance and runtime efficiency, the sampling method choice should be based on specific needs. For a balance between decent performance and high inference throughput, it is advisable to use SAM for point-wise score computation paired with a straightforward sampling policy such as Top-M or a prior-based approach (i.e., SAMBLE with $n_b$=1). This setup delivers good results without significantly impacting speed. On the other hand, if the focus is on achieving an optimal performance on downstream tasks, SAMBLE with the bin-based sampling policy is the ideal choice. However, this method may result in reduced inference throughput due to the complexity of the sampling strategy.

\begin{table}[t]
\centering
\resizebox{\columnwidth}{!}{
\begin{tabular}{@{}lcccccc@{}}
\toprule
Method & \begin{tabular}[c]{@{}c@{}}Attention\\  Map\end{tabular} & \begin{tabular}[c]{@{}c@{}}Sampling\\ Policy\end{tabular} & Params. & FLOPs & \begin{tabular}[c]{@{}c@{}}Throughput\\ (ins./sec.)\end{tabular} & \begin{tabular}[c]{@{}c@{}}OA\\ (\%)\end{tabular} \\ \midrule
FPS + $k$NN & - & - & 20.90k & 2.71G & 102 & 92.80 \\ \midrule
APES (local) & Local & Top-M & 49.15k & 1.09G & 488 & 93.47 \\ \midrule
\multirow{2}{*}{APES (global)} & \multirow{2}{*}{Global} & Top-M & 49.15k & 0.05G & \textbf{520} & 93.81 \\
 &  & Bin-based & 49.92k & 0.38G & 128 & \underline{94.02} \\ \midrule
\multirow{2}{*}{SAMBLE ($n_b=1$)} & \multirow{3}{*}{SAM} & Top-M & 49.15k & 0.05G & \underline{506} & 93.92 \\
 &  & Prior-based & 49.15k & 0.05G & 473 & 93.95 \\ \cmidrule(l){3-7} 
SAMBLE ($n_b=6$) &  & Bin-based & 49.92k & 0.38G & 125 & \textbf{94.18} \\ \bottomrule
\end{tabular}
}
\caption{For model complexity, we report the number of parameters and FLOPs of one downsampling layer for a more detailed comparison. We also report the inference throughput (instances per second) and the classification performance.}
\label{tab:samble_modelComplexity}
\end{table}

\section{More Visualization Results}
\label{sec:moreHistVis}
\begin{figure*}[t]
    \centering
    \includegraphics[width=0.95\linewidth,trim=0 0 0 0,clip]{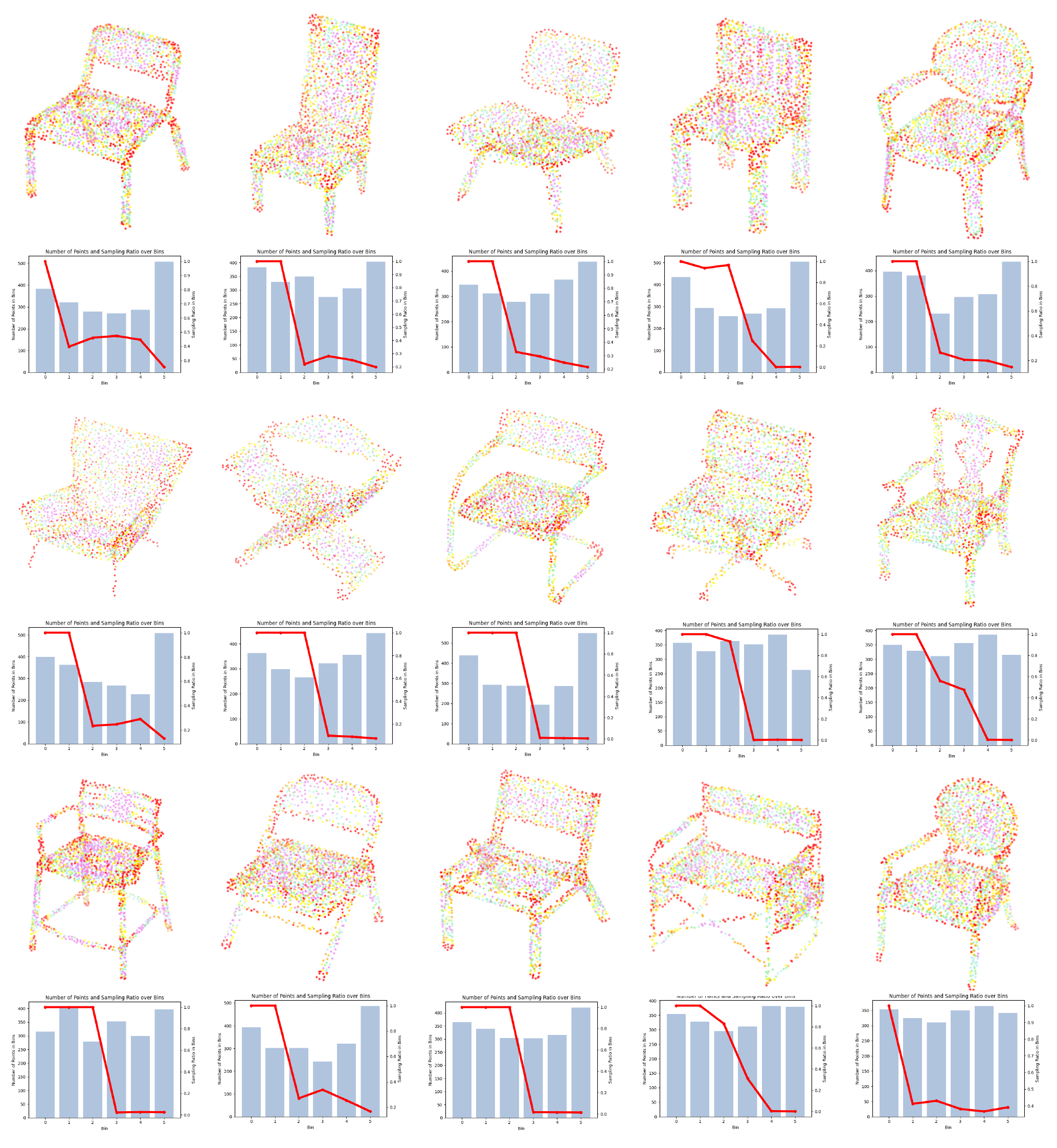}
    \caption{More visualization results of bin partitioning and learned shape-specific sampling strategies on the chair category. Zoom in for optimal visual clarity.}
    \label{fig:moreHist_chair}
\end{figure*}

\begin{figure*}[t]
    \centering
    \includegraphics[width=0.95\linewidth,trim=0 0 0 0,clip]{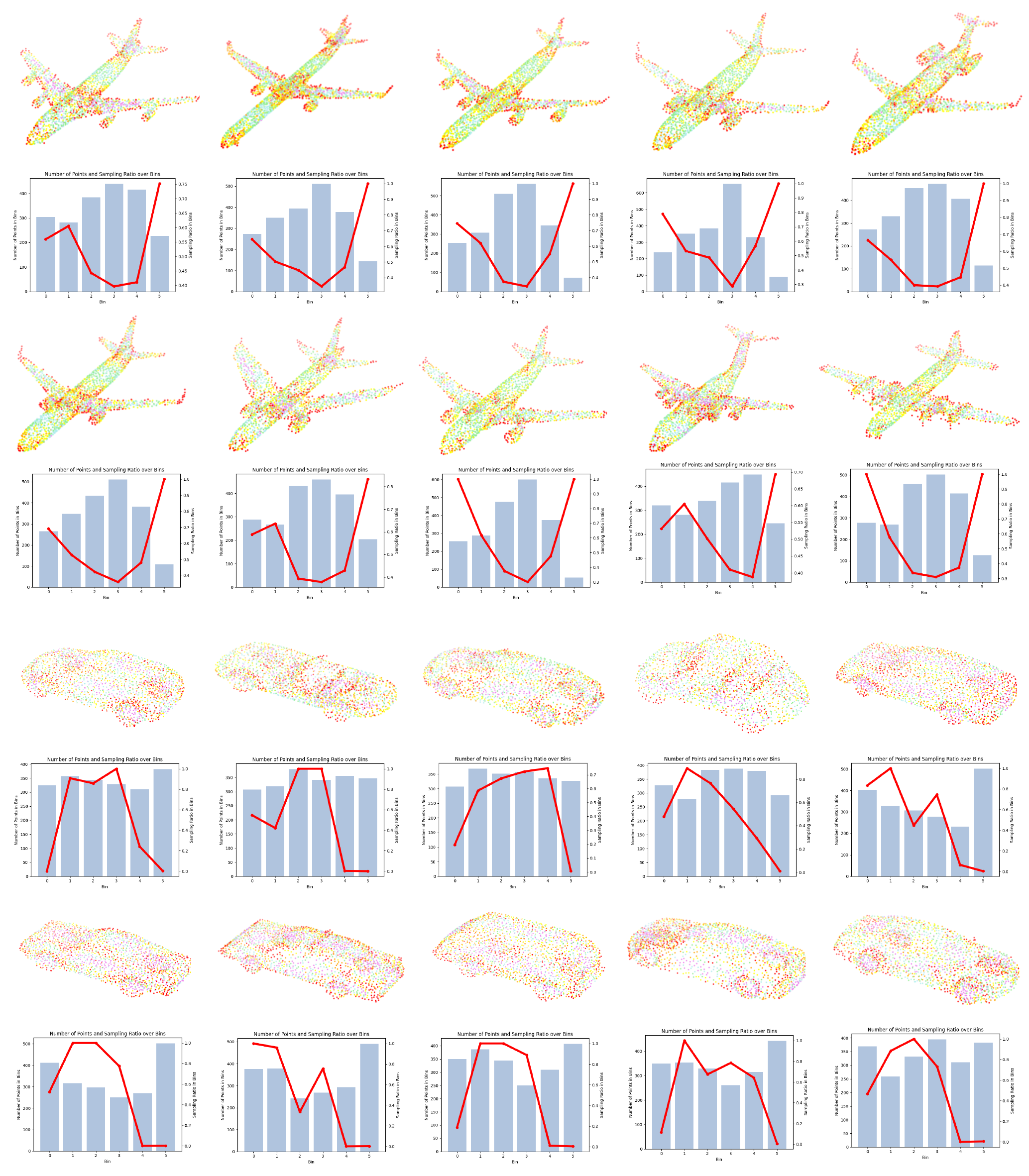}
    \caption{More visualization results of bin partitioning and learned shape-specific sampling strategies on the airplane and car categories. Zoom in for optimal visual clarity.}
    \label{fig:moreHist_planeCar}
\end{figure*}

\begin{figure*}[t]
    \centering
    \includegraphics[width=0.95\linewidth,trim=0 0 0 0,clip]{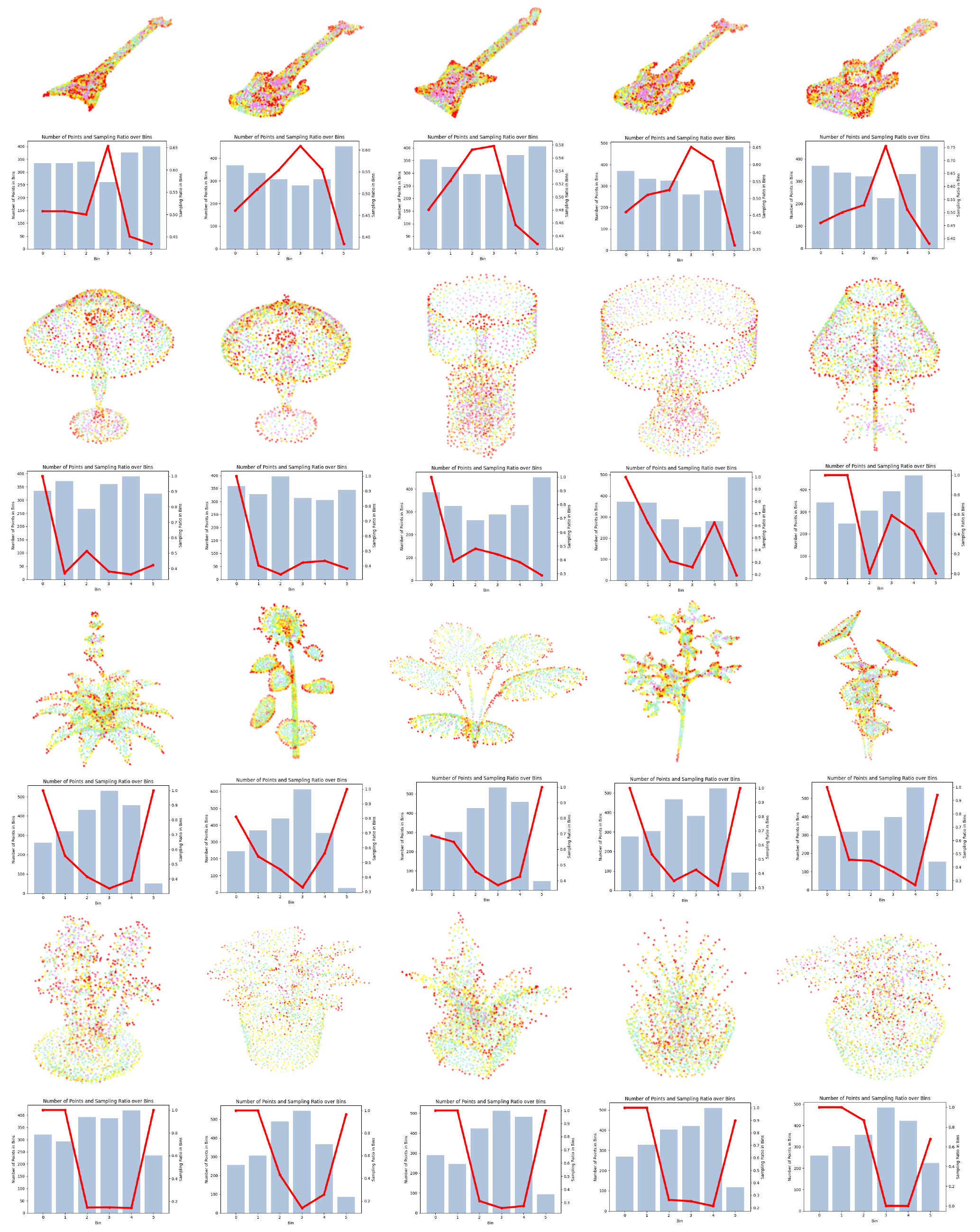}
    \caption{More visualization results of bin partitioning and learned shape-specific sampling strategies on the guitar, lamp, plant, and flower pot categories. Zoom in for optimal visual clarity.}
    \label{fig:moreHist_guitarLampPlantFlowerpot}
\end{figure*}

\begin{figure*}[t]
    \centering
    \includegraphics[width=0.95\linewidth,trim=0 0 0 0,clip]{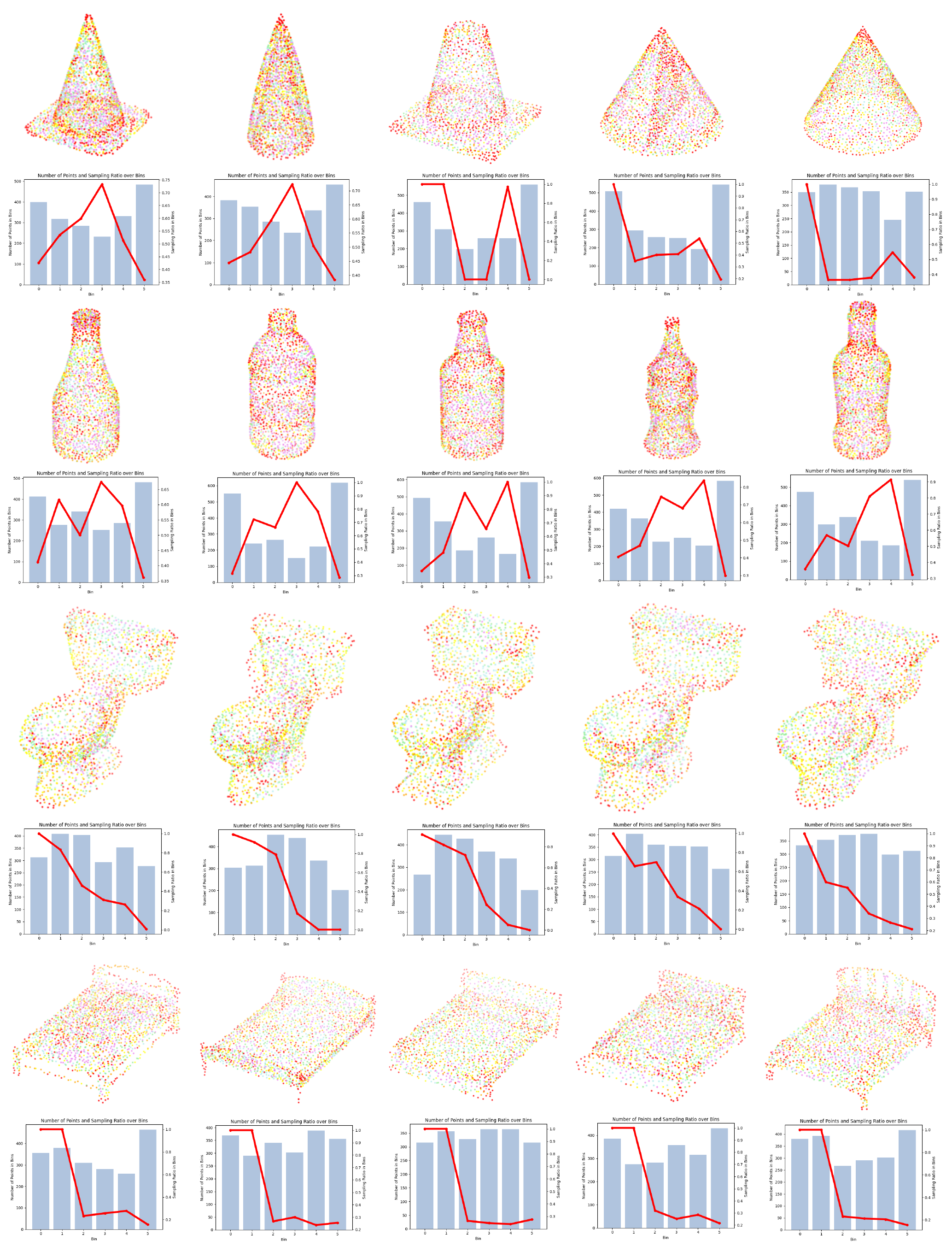}
    \caption{More visualization results of bin partitioning and learned shape-specific sampling strategies on the cone, bottle, toilet, and bed categories. Zoom in for optimal visual clarity.}
    \label{fig:moreHist_coneBottleToiletBed}
\end{figure*}

\textbf{Learned Shape-Specific Sampling Strategies.} 
We present additional extensive results in \cref{fig:moreHist_chair}, \cref{fig:moreHist_planeCar}, \cref{fig:moreHist_guitarLampPlantFlowerpot}, and \cref{fig:moreHist_coneBottleToiletBed} with various categories. 
From them, we can observe that shape edge points are mostly partitioned into the first two bins. Furthermore, in addition to learning shape-wise sampling strategies for individual shapes, it is observed that analogous shapes within the same category exhibit similar histogram distributions and sampling strategies. Conversely, point clouds from different shape categories are sampled by distinct sampling strategies.

\begin{figure*}[t]
    \centering
    \includegraphics[width=0.9\linewidth,trim=0 0 0 0,clip]{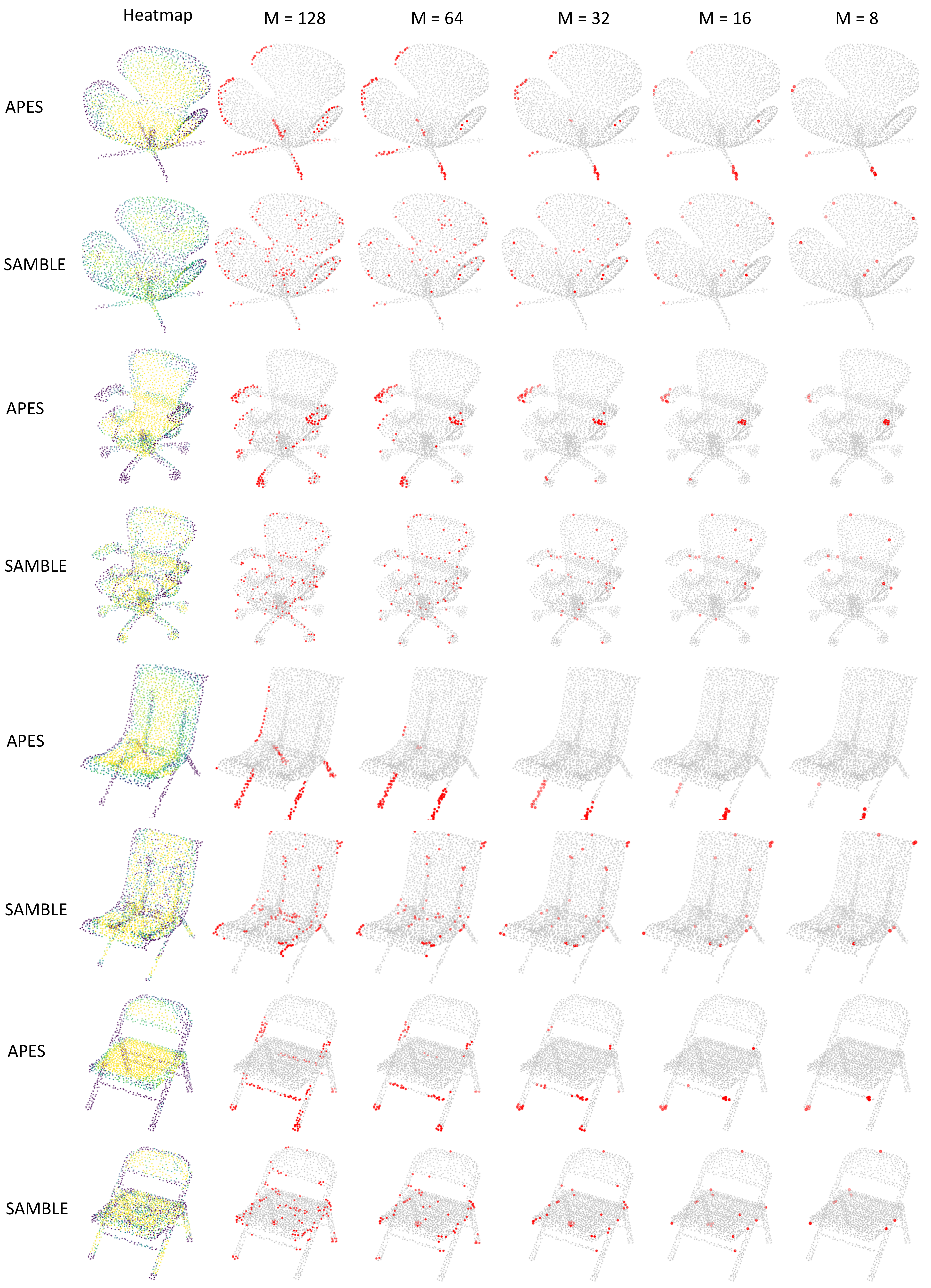}
    \caption{Sampled results of few-point sampling on the chair shapes. No pre-processing with FPS into $2M$ points was performed. Zoom in for optimal visual clarity.}
    \label{fig:moreFew1}
\end{figure*}

\begin{figure*}[t]
    \centering
    \includegraphics[width=0.95\linewidth,trim=0 0 0 0,clip]{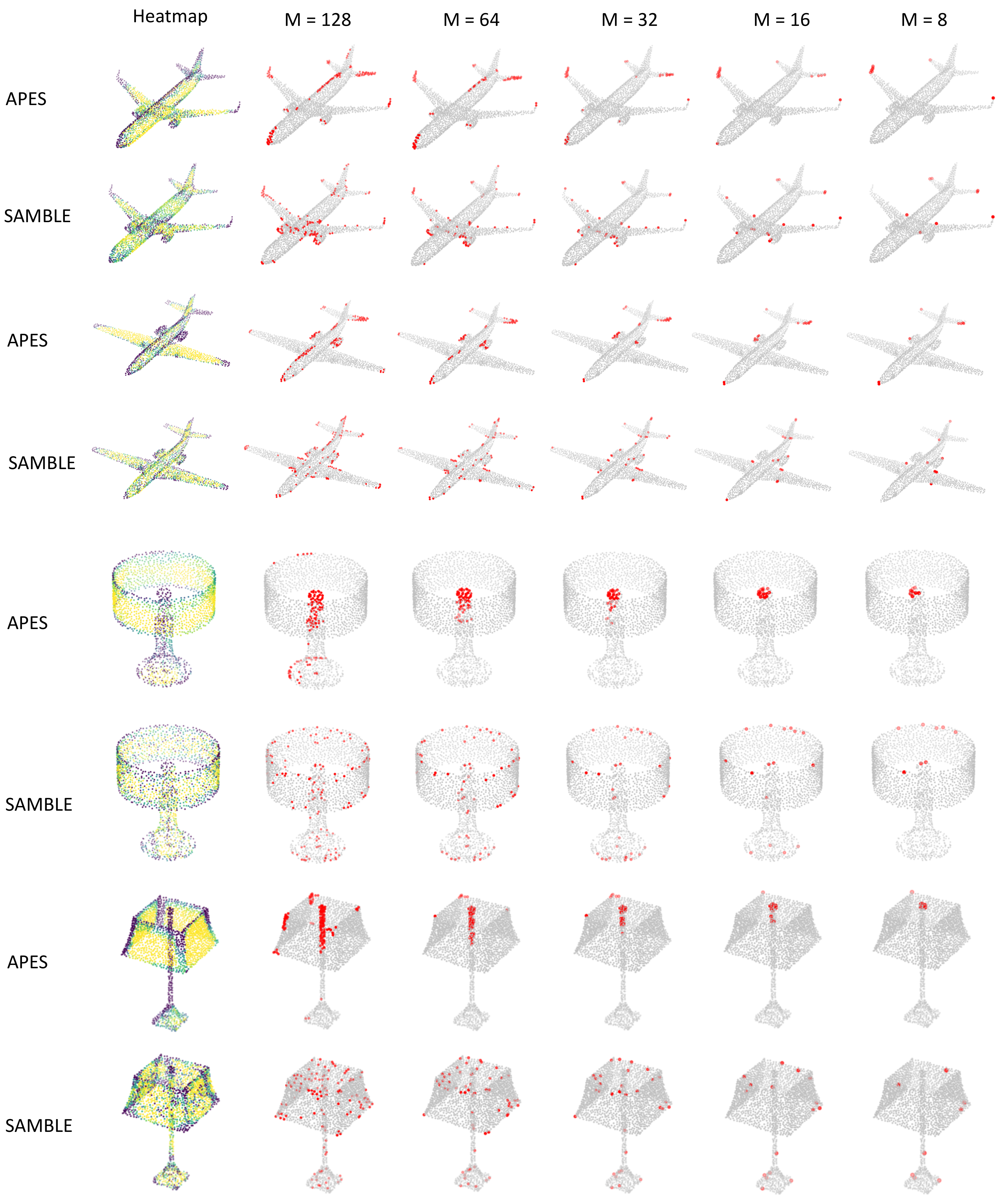}
    \caption{Sampled results of few-point sampling on the airplane and lamp shapes. No pre-processing with FPS into $2M$ points was performed. Zoom in for optimal visual clarity.}
    \label{fig:moreFew2}
\end{figure*}

\vspace{2pt}
\noindent\textbf{Few-Point Sampling.} 
We further provide more visualization results of few-point sampling in \cref{fig:moreFew1} and \cref{fig:moreFew2}. No pre-processing with FPS into $2M$ points was performed.
From them, we can observe that when sampling very few points from the input directly, APES can only sample points from the sharpest regions in a concentrated manner, while our SAMBLE keeps better global uniformity.

\newpage

\end{document}